\documentclass[preprint,times,11pt]{elsarticle}

\biboptions{sort&compress} 

\usepackage{lineno}

\usepackage[utf8]{inputenc}%
\usepackage[T1]{fontenc}
\usepackage{amssymb,amsmath,amsthm,amstext,amscd}%
\usepackage[margin=2.5cm]{geometry}%
 
\usepackage{graphicx}%

\usepackage{todonotes}%
\newcommand{\ADD}[1]{{#1}}%

\usepackage{hyperref}%
\usepackage[ruled, vlined]{algorithm2e}%
\usepackage[capitalise,nameinlink]{cleveref}%
\crefname{figure}{Fig.}{Figs.}%
\crefname{appendix}{}{}%
\crefname{equation}{}{}%
\crefname{alg}{Algorithm}{Algorithms}%
\Crefname{equation}{Equation}{Equations}%
\usepackage{textcomp} %

\usepackage[normalem]{ulem}%
\usepackage{paralist}%
\theoremstyle{plain}%
\newtheorem{theorem}{Theorem}%
\theoremstyle{definition}%
\theoremstyle{remark}%
\newtheorem{remark}[theorem]{Remark}%

\usepackage{bm}%
\usepackage[per-mode=symbol]{siunitx}%
\DeclareSIUnit\Length{L}%
\DeclareSIUnit\Time{T}%
\DeclareSIUnit\liter{l}%
\DeclareSIUnit\molar{M}%

\usepackage{multirow,bigdelim,booktabs,colortbl}%

\DeclareMathOperator{\var}{Var}%
\newcommand{\dd}{\mathrm{d}}%
\newcommand{\rset}{\mathbb{R}}%
\newcommand{\defeq}{\mathrel{\mathop:}=}%
\newcommand{\Deffp}{{D^{\text{eff}}_{+}}}%
\newcommand{\Deffm}{{D^{\text{eff}}_{-}}}%
\newcommand{\keff}{{\kappa^{\text{eff}}}}%
\newcommand{\tp}{{t_{+}}}%
\newcommand{\phiEDL}{{\varphi_{\text{EDL}}}}%
\newcommand{\phiG}{{\varphi_{\Gamma}}}%
\newcommand{\bchi}{\boldsymbol{\chi}}%
\newcommand{\cin}{{c_{\text{in}}}}%
\newcommand{\lpor}{{l_{\text{por}}}}%
\newcommand{\Nsam}{N_{\text{sam}}}%
\newcommand{\Ntrain}{N_{\text{train}}}%
\newcommand{\Ntest}{N_{\text{test}}}%
\newcommand{\unif}{\mathrm{Uniform}}%
\newcommand{\given}{\mid}%
\newcommand{\drop}[1]{{}}%

\makeatletter
\def\ps@pprintTitle{%
 \let\@oddhead\@empty
 \let\@evenhead\@empty
 \def\@oddfoot{}%
 \let\@evenfoot\@oddfoot}
\makeatother
\begin{document}%

\begin{frontmatter}

\graphicspath{
               {./Figures/}
              }

  \title{Mutual Information for Explainable Deep Learning of Multiscale Systems%
  }
    
  \author[first]{S{\o}ren Taverniers\fnref{firstfoot}}%
  \author[second]{Eric
    J.~Hall\corref{mycorrespondingauthor}\fnref{firstfoot}}%
  \cortext[mycorrespondingauthor]{Corresponding author}
  \ead{ehall001@dundee.ac.uk} \author[third]{Markos A.~Katsoulakis}%
  \author[first]{Daniel M.~Tartakovsky\corref{mycorrespondingauthor}}%
  \ead{tartakovsky@stanford.edu} \address[first]{Department of Energy
    Resources Engineering, Stanford University, Stanford, CA 94305,
    USA}%
  \address[second]{Division of Mathematics, University of Dundee,
    Dundee, DD1 4HN, UK}%
  \address[third]{Department of Mathematics and Statistics, University
    of Massachusetts Amherst, Amherst, MA 01003, USA}%
  \fntext[firstfoot]{Both authors contributed equally to this work.}

  \begin{abstract}
Timely completion of design cycles for complex systems ranging from consumer electronics to hypersonic vehicles relies on rapid simulation-based prototyping. The latter typically involves high-dimensional spaces of possibly correlated control variables (CVs) and quantities of interest (QoIs) with non-Gaussian and possibly multimodal distributions. We develop a model-agnostic, moment-independent global sensitivity analysis (GSA) that relies on differential mutual information to rank the effects of CVs on QoIs. \ADD{The data requirements of this information-theoretic approach to GSA are met by replacing  computationally intensive components of the  physics-based model with a deep neural network surrogate. Subsequently, the GSA is used to explain the network predictions, and the surrogate is deployed to close design loops. Viewed as an uncertainty quantification method for interrogating the surrogate, this framework is compatible with a wide variety of black-box models. We demonstrate that the surrogate-driven mutual information GSA provides useful and distinguishable rankings on two applications of interest in energy storage.}
Consequently, our information-theoretic GSA provides  an ``outer loop'' for accelerated product design by identifying the most and least sensitive input directions and performing subsequent optimization over appropriately reduced parameter subspaces. %
  \end{abstract}


\end{frontmatter}

\section{Introduction: GSA and Deep Learning for Simulation-Aided Design }
\label{sec:intro}%

Simulations are a key component of product design as they enable rapid prototyping by guiding costly laboratory tests and investigating regions of the parameter space that are difficult to explore experimentally. To optimize design under uncertainty, an ``outer loop'' can be included to predict the impact of tunable inputs or control variables (CVs) on a system's  quantities of interest (QoIs) \cite{PeherstorferWillcoxGunzburger:2018ol}. In this approach, CVs are treated as random quantities whose distributions are derived from available experimental data, manufacturing constraints, design criteria, engineering judgment, and/or other domain knowledge. Statistical post-processing of repeated solves of a physics-based model for multiple samples of CVs yields the distributions of QoIs. In the context of optimal and robust design and uncertainty quantification (UQ), this outer loop constitutes a \emph{many query} problem that becomes prohibitively expensive when queries rely solely on direct simulation of physics-based models.
 
Data-driven surrogate modeling seeks to alleviate this computational cost by constructing a statistical model for QoIs. Off-the-shelf software such as \texttt{TensorFlow} and \texttt{PyTorch} facilitates the construction of deep learning surrogates, e.g., deep neural networks (DNNs), from data generated by the underlying physics-based model.  This process, which typically involves supervised learning, makes very few assumptions about the nature of the data or data-generating process. This agnosticism makes DNNs suitable for dependent/correlated inputs and non-Gaussian, skewed, multimodal, and/or mutually correlated output QoIs, typically observed in complex real-world systems.  While DNNs (e.g., see \cite{Balokas:2018nn, TripathyBilionis:2018uq, ZhuZabarasEtAl:2019pc, RaissiPerdikarisKarniadakis:2019pinns, HallTaverniersEtAl:2020aa}) can tremendously speed up the design pipeline by accelerating and fully automating the prediction of QoIs, they represent \emph{black boxes} that do not shed any light on the form of the function they are approximating. They provide no clear link between this function and the network weights. Moreover, they are non-identifiable, since two DNNs with the same topology, but different weights, can yield very similar outputs for a given set of input data \cite{GoodfellowEtAl:2016dl}.

Global sensitivity analysis (GSA)~\cite{SaltelliEtAl:2008gs} provides an opportunity to ``peek'' inside a black-box  deep neural networks surrogate  and to interpret its predictions by identifying constellations of input parameters that are likely to yield a targeted model response. GSA facilitates exploration of the entire parameter space and quantifies both first-order (individual) and higher-order (interaction) effects that characterize the contribution of variations in CVs to changes in QoIs. 
Variance-based GSA methods rank input parameters by their contributions to the total variance of a QoI (e.g.,~Sobol' indices \cite{Sobol:1993sa} and total effects \cite{HommaSaltelli:1996im}). 
Their interpretation is ambiguous when spaces of correlated CVs are large~\cite{MaraEtAl:2015np,IoossPrieur:2019se} and QoIs are highly non-Gaussian~\cite{Borgonovo:2006uq,Borgonovo:2007sa}, a situation representative of complex multiscale/multiphysics systems. %
In contrast, moment-independent GSA approaches are easy to interpret regardless of the nature of the data or data-generating process~\cite{ciriello-2019-distribution, CastaingsEtAl:2012sa, VetterTaflanidis:2012sa}; however, they require knowledge of the CV and QoI distributions or availability of sufficient data to approximate them.  Deep neural networks resolve the latter problem by cheaply generating large amounts of data.

A major goal of this study is to harness this synergy between moment-independent GSA and  black-box surrogates and to take advantage of their shared agnosticism to the nature of data and a data-generating process. To this end, we develop an information-theoretic GSA that uses a DNN surrogate to generate sufficient amounts of data. \ADD{Information theory has been used to carry out both local sensitivity analysis using Fisher information matrix methods \cite{MajdaGershgorin:2010qu, KomorowskiEtAl:2011sa, MajdaGershgorin:2011sa, PantazisEtAl:2013rn, PantazisKatsoulakis:2013re} and global sensitivity analysis using mutual information \cite{Critchfield:1986sa,LiuEtAl:2006re,LudtkeEtAl:2008sa,LiuHomma:2009uq,Rahman:2016fs,UmHallEtAl:2019bn}. Furthermore, information inequalities can be used for robust UQ of QoIs in problems with model uncertainty arising, e.g., when models reflect both physics and sparse or multi-sourced data \cite{HallKatsoulakis:2018hd, Feng_Sci_Adv}.}

Our GSA approach utilizes the concept of \emph{differential mutual information} (MI)~\cite{CoverThomas:2006in,Soofi:1994if} to compute Mutual Information Sensitivity Indices (MISIs). \ADD{The latter allow one to interrogate a black-box model by quantifying the dependence of the output QoI with respect to tunable CVs, and to ascertain and rank which CVs are the most influential.} This approach addresses the twin challenges of correlated/dependent CVs and non-Gaussian, skewed, multimodal, and/or mutually correlated QoIs. These features make MISIs an ideal decision-making tool for simulation-aided design.

\ADD{Viewed as an uncertainty quantification method for interrogating surrogates,} our MI-based GSA is compatible with any black-box model such as physics-informed neural networks~\cite{Raissi:2019,RaissiPerdikarisKarniadakis:2019pinns,ZhangLuGuoKarniadakis:2019uq,YangPerdikaris:2019nn,Meng:2020} and ``data-free'' physics-constrained neural networks \cite{Sirignano:2018,Berg:2018,ZhuZabarasEtAl:2019pc,SunEtAl:2020sm}. We leverage a Graph-Informed Neural Network (GINN)~\cite{HallTaverniersEtAl:2020aa} that is tailored for multiscale physics and systems with correlated CVs. The GINN's ability to generate ``big data'' allows us to consider higher-order effects due to interactions between the CVs. In turn, the MI-based rankings help \ADD{to explain} 
the GINN's black-box predictions \ADD{by placing them in context}. %
We validate these rankings by evaluating response curves along sensitive and insensitive directions and comparing these to their counterparts computed with a physics-based model. This comparison provides a clear interpretation of the GINN's predictions in terms of the physics-based model and enables the \ADD{use of the}  GINN to close engineering design loops by deploying it to estimate effect rankings in parameter subspaces yielding optimal QoI values. 

\ADD{ %
Gradient-based methods are often used to interpret DNN predictions. Typically, these methods compute derivatives of the loss function of outputs with respect to inputs using back propagation. Large derivatives with respect to the input, in a suitable metric, are considered influential to the output and, correspondingly, small derivatives are considered less influential. Focusing on a particular class of machine learning tasks related to image classification and pattern recognition and on particular convolutional neural network architectures, these methods use gradients to, e.g., construct saliency maps via local perturbations of the input image \cite{SimonyanVedaldiZisserman:2014sa,AdebayoEtAl:2020fx}, learn importance features through propagation of activation differences \cite{Shrikumar:2017aa}, and construct importance scores that accumulate (integrate) gradients over subsets of perturbations \cite{SundararajanTalyYan:2016gr}.
A %
different approach \cite{ZhangWuZhu:2018nn} is to consider interpretability in terms of the filters of a convolutional neural network by adding an MI term in the loss function in order to retain only mutually independent parts of the DNN during training. In contrast, our work is focuses on applying GSA to scientific machine learning tasks and uses MI (gradient-free) to interrogate surrogate models where the interpretability is understood in relation to the physics-based model. 
}

In the context of multiscale design~\cite{UmHallEtAl:2019bn}, \ADD{we employ} the MISI rankings to interpret the surrogate model's predictions by identifying parameter regions that elicit targeted responses and then using new empirical response data predicted by the GINN for those parameter subspaces to refine an existing prototype. We illustrate how MI-based GSA for explainable DNN predictions enables outer-loop tasks, such as uncertainty quantification and optimal design, to benefit from scientific machine learning. These rankings play a role similar to Shapley values~\cite{StrumbeljKononenko:2011bb}, partial dependence plots~\cite{Friedman:2001gb}, and individual conditional expectation plots~\cite{GoldsteinEtAl:2015bb} found in the statistical learning and data mining literature. Similar to  MISIs, these metrics elucidate the relationship between predicted responses and one or more features in regression models and classifiers based on changes in certain conditional expectations. Unlike these metrics, MISIs depend on distributional---as opposed to moment---information and provide a framework for estimating and ranking higher-order effects, or interactions, among correlated CVs that prove to be crucial for design of complex systems (cf.~\cref{fig:close_design_loop}).

In \cref{sec:misi-development}, we develop a GSA framework that includes both first- and higher-order MISIs. In \cref{sec:ginns}, this methodology is combined with a GINN in the context of a testbed problem related to the design of a supercapacitor. Validation of the MI-based rankings and closure of design loops through subsequent rankings in the reduced parameter space with optimal QoI values are performed in \cref{sec:misi-validation}. In \cref{sec:conclusions}, we summarize the main conclusions drawn from this study and discuss future work.

\section{MISIs for Model-Agnostic GSA}
\label{sec:misi-development}

We consider a model $\mathcal{M}$ of a complex physical system,
\begin{equation}
  \label{eq:surrogate-model}
  \bm{Y} = \mathcal{M}(\bm{X}) \,,
\end{equation}
that predicts the response of a collection of QoIs $\bm{Y} \in \rset^q$ to a collection of tunable CVs $ \bm{X} \in \rset^p$. The model $\mathcal{M}$ propagates distributions on CVs to distributions on QoIs query-by-query, i.e., it generates one sample response $\bm{Y}^{(m)}$, $m=1,\dots,M$, for each input sample or observation $\bm{X}^{(m)}$. We 
place no restrictions on the dependence structure of the CVs or on the nature of their functional relationship to the response. 

\subsection{First-order effects described by differential MI}%
\label{sec:first-order-effects}%

Differential MI is a pseudo-distance used in machine learning \cite{Bishop:2006ml,KollerFriedman:2009gm} and model selection \cite{BurnhamAnderson:2002ms}, among others. %
Quantifying the amount of shared information between $\bm{V} \in \rset^{d_1}$ and $\bm{W} \in \rset^{d_2}$, the differential MI is defined as~\cite{CoverThomas:2006in},
\begin{equation}
  \label{eq:mi}
  I(\bm{V}; \bm{W})
  \defeq \iint_{\mathcal{V} \otimes \mathcal{W}}
  \log\left( \frac{f_{\bm{V},\bm{W}}(\bm{v}, \bm{w})}{f_{\bm{V}}(\bm{v})f_{\bm{W}}(\bm{w})} \right)
  f_{\bm{V},\bm{W}}(\bm{v},\bm{w}) \dd{\bm{v}} \dd{\bm{w}}\,,
\end{equation}
where $f_{\bm{V}}$, $f_{\bm{W}}$, and $f_{\bm{V},\bm{W}}$ denote marginal and joint probability density functions (PDFs) with support $\mathcal{V}$, $\mathcal{W}$, and $\mathcal{V} \otimes \mathcal{W}$, respectively. 
The differential MI possesses many of the same properties as the discrete MI, including symmetry $I(\bm{V}; \bm{W}) = I(\bm{W}; \bm{V})$ and non-negativity $I(\bm{V}; \bm{W}) \geq 0$ (with equality if and only if $\bm{V}$ and $\bm{W}$ are independent). Unlike its discrete counterpart, the differential MI can take on infinite values, e.g., if $\bm{V} = \bm{W}$. The following features make the differential MI appropriate for GSA in multiscale design:
\begin{compactenum}[\quad(i)]
\item its interpretation does not rely on the dependence structure of the CVs,
\item its moment independence makes it suitable for a wide range of CV and QoI PDFs, and
\item its continuous nature is suitable for analysis of continuous systems.
\end{compactenum}
The first two features enable a model-agnostic implementation, while the last one facilitates uncertainty quantification for downstream computations relying on continuous QoIs. 

To describe the first-order effect of a CV $X \in \bm{X}$ on a target QoI $Y \in \bm{Y}$, we define a MISI,
\begin{equation}
  \label{eq:misi}
  S_{Y}(X) \defeq I(X; Y),
\end{equation}
and interpret it as a measure of the strength of the association between $X$ and $Y$. A large score indicates that $X$ is a globally influential CV in the design of $Y$ relative to the PDF of $\bm{X}$. In complex systems, $Y$ is unlikely to be completely described by a single CV $X$, so the value of $S_Y(X)$ in \cref{eq:misi} is likely to remain finite. Since $S_Y(X)$ places equal importance on linear and nonlinear relationships due to the self-equitability of the MI \cite{KinneyAtwal:2014pn}, it recovers the rankings of Sobol' indices in the setting of independent CVs $\bm X$, i.e., when Sobol' rankings are justified. %

The MISI $S_Y(X)$ in \cref{eq:misi} can be estimated from empirical data generated by querying the model $\mathcal{M}$. A plug-in Monte Carlo estimator for $S_Y(X)$,
\begin{equation}
  \label{eq:misi-est}
  \widehat{S}_Y(X) \defeq \frac{1}{M} \sum_{m=1}^{M} \log \left(\frac{\widehat{f}_{X,Y}(X^{(m)},Y^{(m)}; b_{X},b_{Y})}
    {\widehat{f}_{X}(X^{(m)}; b_{X})\widehat{f}_{Y}(Y^{(m)}; b_{Y})}\right), 
\end{equation}
can be computed via joint and marginal kernel density estimators (KDEs) $\widehat{f}$ at input-output data pairs $\{X^{(m)}\,, Y^{(m)}\}$, $m=1,\dots, M$~\cite{KrishnamurthyEtAl:2014rd,KandasamyEtAl:2015vm}. %
A Gaussian kernel KDE $\widehat{f}_{\bm{Z}}$ for an unknown PDF $f_{\bm{Z}}$ based on $M'$  identically distributed observations $\bm{Z}^{(1)}, \dots, \bm{Z}^{(M')}$ of $\bm{Z} \in \rset^d$ is given by~\cite{Wasserman:2006np}
\begin{equation}
  \label{eq:kde_Gaussian_diag}
  \widehat{f}_{\bm{Z}}(\boldsymbol{z}; \mathbf{b}) %
  \defeq \frac{(2\pi)^{-d/2}}{M' \prod_{j=1}^d b_j} \sum_{m=1}^{M'}
  \prod_{j=1}^d \exp \left[ -\frac{\left(z_j-Z_j^{(m)}\right)^2}{2b_j^2} \right], \quad \bm{z} \in \rset^d.
\end{equation}
Among many algorithms for the automated computation of the positive bandwidth parameters $\mathbf{b}=(b_1,\dots,b_d)^\top$, we chose a direct plug-in bandwidth selector called the improved Sheather--Jones method \cite{BotevGrotowskiKroese:2010kd}. To ensure that the joint and marginal KDEs in \cref{eq:misi-est} are defined consistently, i.e., that
\begin{equation*}
  \int_{\mathcal{X}} \widehat{f}_{X,Y}(x,y; b_{X},b_{Y}) \dd{x}
  = \widehat{f}_{Y}(y;b_{Y})
  \quad\text{and}\quad
  \int_{\mathcal{Y}} \widehat{f}_{X,Y}(x,y; b_{X},b_{Y}) \dd{y}
  = \widehat{f}_{X}(x;b_{X})\,,
\end{equation*}
we require the smoothing bandwidths for the joint and marginal PDFs to be equal. 
That is, the bandwidths related to $X$ in $\widehat{f}_{X}$ and $\widehat{f}_{X,Y}$ must be the same.

The KDE-based direct plug-in estimator $\widehat{S}_Y(X)$ in~\cref{eq:misi-est} is %
easy to implement. %
However, its computation %
is not sample-efficient and, hence, unfeasible in the absence of an efficient surrogate; moreover, KDEs are anticipated to fail in high dimensions \cite{KrishnamurthyEtAl:2014rd}. In such circumstances, one can deploy alternative strategies for estimating MI, such as a non-parametric $k$-nearest neighbor algorithm~\cite{KraskovEtAl:2004mi} and a non-parametric neural estimation approach suitable for high-dimensional PDFs~\cite{BelghaziEtAl:2018mi}. Contrary to the discrete MI indices~\cite{Critchfield:1986sa,LudtkeEtAl:2008sa}, non-parametric density estimators, such as~\cref{eq:misi-est}, introduce no  bias associated with a quantization of the QoIs, whose continuous nature may need to be preserved for the purpose of a downstream computations.

\begin{remark}[Independent input-output]
  The plug-in estimator \cref{eq:misi-est} involves the joint PDFs, i.e., is useful when input-output sample pairs are available. A change of measure in \cref{eq:mi} yields an equivalent estimator,
  \begin{equation}
    \label{eq:misi-indep-io-est}
    \widehat{S}^{\perp}_Y(X) \defeq \frac{1}{M} \sum_{m=1}^{M} \log \left(\frac{\widehat{f}_{X,Y}(X^{(m)},Y^{(m)}; b_{X},b_{Y})}
      {\widehat{f}_{X}(X^{(m)}; b_{X})\widehat{f}_{Y}(Y^{(m)}; b_{Y})}\right)
    \frac{\widehat{f}_{X,Y}(X^{(m)},Y^{(m)}; b_{X},b_{Y})}
    {\widehat{f}_{X}(X^{(m)}; b_{X})\widehat{f}_{Y}(Y^{(m)}; b_{Y})}\,,
  \end{equation}
 that is suitable for independent samples from the input and output distributions \ADD{as the  expectation is no longer with respect to the joint PDF but with respect to each marginal PDF} (cf.~\cite{UmHallEtAl:2019bn}).
\end{remark}

\subsection{Higher-order effects described by conditional differential
  MI}
\label{sec:high-order-effects}

For large spaces of possibly correlated CVs, %
it is of interest to also consider the impact of interactions among subsets of CVs on a given QoI. To describe the effects of pairwise interactions between $X_{1}, X_{2} \in \bm{X}$ on a target QoI $Y$, we define a second-order MISI,
\begin{equation}
  \label{eq:misi-2nd}
  S_Y(X_1,X_2) \defeq I(X_1; X_2 \given Y),
\end{equation}
in terms of the conditional differential MI,
\begin{equation}
  \label{eq:cond-mi}
  I(\bm{V}; \bm{W} \given \bm{U}) \defeq
  \iiint_{\mathcal{U}\otimes
    \mathcal{V} \otimes \mathcal{W}}
  \log\left(
    \frac{f(\bm{u}) f(\bm{u}, \bm{v}, \bm{w})}
    {f(\bm{u}, \bm{w}) f(\bm{u}, \bm{v})} \right) f (\bm{u}, \bm{v}, \bm{w})
  \dd{\bm{w}} \dd{\bm{v}} \dd{\bm{u}}.
\end{equation}
The latter represents the MI between $\bm{V}$ and $\bm{W}$ conditioned on $\bm{U}$ that we express  in terms of joint and marginal PDFs.\footnote{Here and in the sequel we suppress the labels on densities when the distribution is clear from the context.} The conditional MI in~\cref{eq:cond-mi} is related to the MI in~\cref{eq:mi} through the chain rule,
\begin{equation}
  \label{eq:mi-chain-rule}
  I(V_1,V_2,\dots,V_{k};W) = \sum_{i=1}^{k}I(V_i;W|V_{i-1},V_{i-2},\dots,V_1),
\end{equation}
for $V_1,V_2,\dots, V_{k}\in \bm{V}$ and $W\in\bm{W}$ where zero-indexed sets in the conditioning are empty.
To see that \cref{eq:misi-2nd} captures only the second-order effects, we note that $I(X_{1}, X_{2}; Y)$ 
describes the full effect of the pair $(X_{1},X_{2})$ on $Y$. According to~\cref{eq:mi-chain-rule}, the full second-order effect is expressed as
\begin{equation}
  \label{eq:full-2nd-effect}
  I(X_{1},X_{2}; Y) = I(X_{1}; Y) + I(X_{2}; Y)
  - I(X_{1};X_{2}) + I(X_{1};X_{2} \given Y),
\end{equation}
which includes first-order effects $I(X_{1};Y)$ and $I(X_{2};Y)$, while $I(X_{1};X_{2})$ captures the interaction between $X_{1}$ and $X_{2}$ (the latter term vanishes if $X_{1}$ and $X_{2}$ are independent). The remaining conditional differential MI in \cref{eq:full-2nd-effect} describes the desired second-order effect. 

A plug-in Monte Carlo estimator for the second-order index \cref{eq:misi-2nd} is 
\begin{equation}
  \label{eq:misi-2nd-est}
  \widehat{S}_Y(X_1,X_2) \defeq
  \frac{1}{M} \sum_{m=1}^{M}
  \log \left(\frac{\widehat{f}\drop{_{Y_j}}(Y^{(m)}; b_{Y})
      \widehat{f}\drop{_{X_{1},X_{2},Y}}(X_{1}^{(m)},X_{2}^{(m)},Y^{(m)}; b_{X_1},b_{X_2},b_{Y})}
    {\widehat{f}\drop{_{X_{1},Y}}(X_{1}^{(m)},Y^{(m)}; b_{X_1},b_{Y})
      \widehat{f}\drop{_{X_{2},Y}}(X_{2}^{(m)},Y^{(m)}; b_{X_2},b_{Y})}\right),
\end{equation}
based on input-output triples $(X_1^{(m)}, X_2^{(m)}, Y^{(m)})$, $m=1,\dots,M$. The plug-in estimator is justified in the context of surrogate modeling and is easy to implement using KDEs \cref{eq:kde_Gaussian_diag} with suitably equalized bandwidths. It can be built from the same sample data used to evaluate \cref{eq:misi-2nd-est} and comes with the same caveats.

A $k$th-order MISI (with $k>2$) is defined as
\begin{equation}
  \label{eq:misi-kth}
  S_Y(X_1,\dots, X_k) \defeq I(X_1; X_2; \dots ; X_k \given Y).
\end{equation}
It quantifies the impact of the interaction among the collection of variables $\{X_1, \dots, X_k\} \subset \bm{X}$ on $Y$. The conditional multivariate differential MI is defined inductively,
\begin{equation}
  \label{eq:cond-mv-diff-mi}
  I(X_1; X_2; \dots ; X_k \given Y)
  =  I(X_2;\dots ; X_k \given X_1, Y) - I(X_2;\dots ; X_k \given Y).
\end{equation}
It is symmetric with respect to permutation of the variables $X_j$, with $1\leq j\leq k$. For example, the third-order effect of the interactions among the triple $X_{1}$, $X_{2}$, and $X_{3}$ on $Y$ is given by the third-order MISI,
\begin{equation}
  \label{eq:misi-3rd}
  S_Y(X_1,X_2,X_3)
  = I(X_{1}; X_{2}; X_{3} \given Y)
  =  I(X_2; X_3 | X_1, Y) - I(X_2; X_3 \given Y).
\end{equation}
The conditional multivariate MI in~\cref{eq:cond-mv-diff-mi} and, hence, $S_Y$ in \cref{eq:misi-kth} can be either positive or negative. They are related to a conditional form of the ``interaction information''~\cite{McGill:1954mi} and the ``co-information''~\cite{Bell:2003mi}. Instead of interpreting  such higher-order ($k > 2$) effects, we focus on algorithms for ranking the first- and second-order effects with appropriate confidence intervals (cf.~\cref{fig:ranking-1st} for first-order and \cref{fig:ranking-2nd} for second-order effect rankings).

\subsection{Algorithms for MISI ranking with confidence}
\label{sec:ranking-effects}%

We %
assume the availability of a surrogate model $\mathcal{M}$ for generating large amounts of response data. While based on slightly different theoretical approaches, the two algorithms described below enable the constructing of first- and higher-order effect rankings for a given QoI with a focus on the generation of associated confidence intervals. The latter enable distinguishing closely-ranked pairs of CVs.

\subsubsection{\cref{alg:rank-cis}: Compute MISIs with confidence intervals adjusted for ranking}

\cref{alg:rank-cis} (see the pseudocode) constructs the plug-in estimators $\widehat{S}_Y(X_j)$ in \cref{eq:misi-est} with confidence intervals selected such that pairwise comparisons of the MISIs and their accompanying intervals determine the effect ranks. For fixed $Y$, we order the $\widehat{S}_Y(X_j)$ as
\begin{equation}
  \widehat{S}_Y(X_{j_1}) > \widehat{S}_Y(X_{j_2}) > \cdots > \widehat{S}_Y(X_{j_p}).
\end{equation}
Then, the ranked first-order MISI estimators are
\begin{equation}
  \label{eq:ranking-1st}
  (\widehat{\theta}_1, \dots, \widehat{\theta}_p) \defeq
  \left(\widehat{S}_Y(X_{j_1}) , \dots , \widehat{S}_Y(X_{j_p})  \right),
\end{equation}
where the individual (additive) effects of the CVs are arranged in order of importance, from the greatest ($\widehat{\theta}_1$) to the least ($\widehat{\theta}_p$). The \ADD{(true)} rank $r_j$ of CV $X_j$ is estimated by the plug-in quantity,
\begin{equation}
  \label{eq:rank}
  \widehat{r}_j
  \defeq p - \#\{ \widehat{S}_Y(X_i) < \widehat{S}_Y(X_j),\quad
  i=1,\dots,p\},
\end{equation}
\ADD{where $\#\{\cdot\}$ denotes the cardinality of the set}, such that
\begin{equation}
  \label{eq:theta-rank-estimator}
  \widehat{\theta}_{\,\widehat{r}_j} = \widehat{S}_Y(X_j).
\end{equation}
Since the MISI $S_Y(X)$ in~\cref{eq:misi} is a global measure of sensitivity, \cref{eq:ranking-1st} represents a global ranking of the first-order effect of each CV relative to the distribution of $\bm{X}$.

\begin{algorithm}[!ht]
  \DontPrintSemicolon \SetKw{KwSort}{sort}\SetKw{KwBreak}{break}%
  \SetKw{KwBy}{by}%
  \SetKw{KwSet}{set}
  \SetKwInOut{Input}{input}\SetKwInOut{Output}{output}
  \SetKwComment{Comment}{$\triangleright$\ }{}%

  \Input{$\mathcal{M}$ \Comment*{Surrogate model
      \cref{eq:surrogate-model}}}%
  \Input{$(X^{(1)}_1, \dots, X^{(1)}_p), \dots, (X^{(M)}_1, \dots,
    X^{(M)}_p)$ \Comment*{$M$ independent samples of CVs}}%
  \Input{$\bar{\gamma}$, $\mathrm{TOL}$ \Comment*{Non-overlap sig.\
      and tolerance}}%
  \BlankLine %
  \Output{$(\widehat{\theta}_1 \pm z[\beta/2]
    \widehat{\sigma}_1 \,, \dots\,, \widehat{\theta}_p \pm
    z[\beta/2] \widehat{\sigma}_p )$ \Comment*{Rankings with
      adjusted intervals \cref{eq:rank-1st-ci-adjusted}}}%
  \BlankLine %
  \Begin{%
    \emph{Compute target QoI observations}\;%
    \For{$m \in \{1, \dots, M\}$}{
      $Y^{(m)} \leftarrow \mathcal{M}(X^{(m)}_1, \dots, X^{(m)}_p)$}%
    \BlankLine %
    \emph{Compute and rank first-order MISI estimators}\;%
    \For{$j \in \{1, \dots, p\}$}{%
      $G(x,y) \leftarrow \log \bigl[ \widehat{f}_{X_j,Y}(x, y) \bigr]
      - \log \bigl[\widehat{f}_{X_j}(x) \widehat{f}_Y(y)\bigr] $ \;%
      $\widehat{S}_Y(X_j) \leftarrow \frac{1}{M} \sum_{1 \leq m \leq
        M} G(X_j^{(m)}, Y^{(m)})$ \Comment*{First-order MISI
        \cref{eq:misi-est}}%
    }%
 
    \For{$j \in \{ 1, \dots, p\}$}{%
      $\widehat{r}_{j} \leftarrow p - \#\{ \widehat{S}_Y(X_i) <
      \widehat{S}_Y(X_j)\,,\; i=1,\dots,p\}$ \Comment*{Rank of $j$th
        CV \cref{eq:rank}}%
      $\widehat{\theta}_{\,\widehat{r}_j} \leftarrow \widehat{S}_Y(X_j)$
      \Comment*{Ranked MISI \cref{eq:theta-rank-estimator}}
      $\widehat{\sigma}_{\, \widehat{r}_j} \leftarrow \bigl( \var
      [\widehat{S}_Y(X_j)]\bigr)^{1/2}$ \Comment*{MISI standard error
        \cref{eq:vanilla-ci}}} 
  \BlankLine %
  \emph{Compute comparison-adjusted confidence intervals with 
  average type I error $\bar{\gamma}$}\;%
  \For{$k,l \in \{ 1, \dots, p\}$}{%
    $s_{kl} \leftarrow (\widehat{\sigma}_k + \widehat{\sigma}_l)/
    (\widehat{\sigma}_k^2 + \widehat{\sigma}_l^2)^{1/2}$\;%
  }%
  $f(z) \leftarrow \bar{\gamma} - \frac{4}{p(p-1)} \sum_{1\leq k<l\leq
    p} (1-\Phi(z s_{kl}))$ \Comment*{Average non-overlap sig.\
    \cref{eq:type-1-error-averaged}}%
  $f'(z) \leftarrow \frac{4}{p(p-1)} \sum_{1\leq k<l\leq p} \varphi(z
  s_{kl}) \cdot s_{kl}$ \Comment*{Normal PDF $\varphi = \Phi'$}%
  $z_0 \leftarrow \Phi^{-1}(1 - \bar{\gamma}/2) \cdot
  (\widehat{\sigma}_1^2 + \widehat{\sigma}_2^2)^{1/2} /
  (\widehat{\sigma}_1 + \widehat{\sigma}_2)$ \;%
  $i \leftarrow 0$\;
  \While{$\mathrm{err}_i < \mathrm{TOL}$}{
    $z_{i+1} \leftarrow z_{i} - f(z_i) / f'(z_i)$ \Comment*{Newton--Raphson iterations}%
    $\mathrm{err}_i \leftarrow | z_{i+1} - z_i | / z_i$ \;%
    $i \leftarrow i + 1\;$
  }%
  $z[\beta/2] \leftarrow z_{i+1}$ \Comment*{Difference level
    \cref{eq:rank-1st-ci-adjusted}}%
  \BlankLine %
  \Return{$(\widehat{\theta}_1, \dots, \widehat{\theta}_p)$,
    $(\widehat{\sigma}_1, \dots, \widehat{\sigma}_p)$,
    $z[\beta / 2]$}%
}%
\caption{Compute first-order MISIs with confidence intervals adjusted for ranking}%
\label{alg:rank-cis}
\end{algorithm}

One could approximate the standard $100 \cdot (1-\alpha)\%$ confidence interval for $\theta_k$ with
\begin{equation}
  \label{eq:vanilla-ci}
  (\widehat{\theta}_k - z[\alpha/2] \widehat{\sigma}_k, \quad
    \widehat{\theta}_k + z[\alpha/2] \widehat{\sigma}_k ), \qquad 
     \widehat{\sigma}_{k} \defeq \sqrt{\var \bigl[\widehat{\theta}_k\bigr]}
  = \sqrt{ \var \bigl[ \widehat{S}_Y(X_{j_k}) \bigr]},
\end{equation}
where $\Phi(z[\alpha/2]) = 1- \alpha/2$ for the standard normal cumulative distribution function $\Phi$, and $\widehat{\sigma}_{k}$
is the standard error.\footnote{\ADD{Here, $z[\alpha]$ denotes the value such that $\Phi(z[\alpha]) = 1- \alpha$, i.e., the value such that the standard normal PDF in the right-hand tail is equal to $\alpha$.}} However, this confidence interval does not readily distinguish the rankings. The fact that the $100\cdot(1-\alpha)\%$ confidence intervals for the ranked effects $\theta_k$ and $\theta_l$, with $k\neq l$, fail to overlap does not necessarily mean that the difference in the rankings is statistically significant at the $\alpha$ level. Following~\cite{GoldsteinHealy:1995re},  \cref{alg:rank-cis} reports confidence intervals with comparison-adjusted widths, such that the non-overlap significance level meets a given threshold on average.
Assuming normality and independence of $\widehat{\theta}_k$ and $\widehat{\theta}_l$, the confidence intervals at level $\beta$ do not overlap if
\begin{equation}
  \label{eq:rank-difference}
  |\widehat{\theta}_k - \widehat{\theta}_l| >
  z[\beta/2] (\sigma_k + \sigma_l).
\end{equation}
Inequality~\cref{eq:rank-difference} holds with probability $1-\gamma_{kl}$, where the pairwise non-overlap significance level $\gamma_{kl}$ is given by
\begin{equation}
  \label{eq:type-1-error}
  \gamma_{kl} \defeq 2 - 2 \Phi\Bigl(z[\beta/2]
    \frac{\sigma_k + \sigma_l}
    {\sigma_{kl}}\Bigr),
  \qquad \sigma_{kl}
  \defeq \sqrt{\sigma_k^2 + \sigma_l^2}.
\end{equation}
We select the level $\beta$ to ensure that the average of the pairwise errors $\gamma_{kl}$ over all $1 \leq k < l \leq p$ is at a predefined level $\bar{\gamma}$ (set to $\bar{\gamma} = 0.01$), i.e.,
\begin{equation}
  \label{eq:type-1-error-averaged}
  \bar{\gamma} - \frac{2}{p(p-1)} \sum_{1 \leq k < l \leq p} \gamma_{kl}  = 0.
\end{equation}
The level $\beta$, for which \cref{eq:type-1-error-averaged} holds, is found via Newton--Raphson iteration (cf. \cref{alg:rank-cis}) using sample estimates for the standard errors. For each ranked effect $\theta_k$, the approximate confidence interval at level $\beta$ is
\begin{equation}
  \label{eq:rank-1st-ci-adjusted}
  \left(\widehat{\theta}_k - z[\beta / 2] \widehat{\sigma}_k, \;
    \widehat{\theta}_k + z[\beta / 2] \widehat{\sigma}_k \right), \qquad
  \quad 1\leq k \leq p;
\end{equation}
the error bars indicate that the non-overlap significance level is $\bar{\gamma}$ on average. The intervals \cref{eq:rank-1st-ci-adjusted} provide the visual comparison of pairwise effects with clear interpretation: overlapping/non-overlapping intervals imply that the associated ranks are indistinguishable/distinguishable.

With a slight modification, \cref{alg:rank-cis} can be used to rank the second-order MISI estimators~\cref{eq:misi-2nd},
\begin{equation}
  \label{eq:ranking-2nd} (\widehat{\zeta}_1, \dots,
\widehat{\zeta}_{p'}) \defeq \left(\widehat{S}_Y(X_{j_1},X_{k_1}) ,
\dots , \widehat{S}_Y(X_{j_{p'}},X_{k_{p'}}) \right),
\end{equation}
where $p' \defeq p!/(2!(p-2)!)$ is the number of pairs $(X_j,X_k)$ with $j<k$ that can be formed from $\bm{X}$. %
The computation of \cref{eq:ranking-2nd} replaces that of the first-order indices in \cref{alg:rank-cis}; the computation of the comparison-adjusted confidence intervals for $\zeta_i$, $i=1, \dots, p'$, proceeds analogously to that of the first-order confidence intervals~\cref{eq:rank-1st-ci-adjusted} with the non-overlap significance averaged over all pairs $(\widehat{\zeta}_k,\widehat{\zeta}_l)$ with $1 \leq k \leq l \leq p'$.

\subsubsection{\cref{alg:rank-dist-method}: Rank MISIs with percentile confidence intervals}

\cref{alg:rank-dist-method} (see the pseudocode) constructs non-parametric estimates with confidence intervals for the unknown rankings $r_k$ directly from the sampling distribution. In contrast to \cref{alg:rank-cis} which uses normality theory, the present method builds a distribution for each rank by repeated observation of the MISIs using the surrogate model. For each $X_j\in\bm{X}$, we compute MISI replications,
\begin{equation}
  \label{eq:nth-rep-misi}
  \widehat{s}^{\,(n)}_j \defeq \widehat{S}_{Y}^{\,(n)}(X_j)\,,
  \quad n = 1, \dots, N\,,
\end{equation}
 from input-output sample pairs $(X_j^{(m)},Y^{(m)})$, $m=1, \dots, M$. That is, for each replication, we generate $M$ new input-output pairs $(\bm{X}, Y)$ using the surrogate model and compute \cref{eq:nth-rep-misi} for every $j = 1, \dots, p$ with these observations. From these replications, we use~\cref{eq:rank} to compute rank replications $\widehat{r}_k^{\,(n)}$, $k=1,\dots,p$. The rank estimators are
\begin{equation}
  \label{eq:rank-est-1st}
  \widehat{r}_k = \frac{1}{N} \sum_{n=1}^N \widehat{r}^{\,(n)}_k,
\end{equation}
and the corresponding $\delta$ percentile confidence intervals are
\begin{equation}
  \label{eq:rank-est-1st-ci}
  (\widehat{r}_k[\delta/2], \;  \widehat{r}_k[1-\delta / 2]),
  \qquad 1 \leq k \leq p.
\end{equation}
The equal-tail percentiles $\widehat{r}_k[\delta/2]$ and $\widehat{r}_k[1- \delta / 2]$ are estimated from the replications in the spirit of the bootstrap percentile confidence intervals~\cite{Efron:1981ci}.

The computational burden of \cref{alg:rank-dist-method} is greater than that of \cref{alg:rank-cis}, since the estimator is computed for each of the $N$ replications. Yet, this method is non-parametric and its results are anticipated to be more easily interpretable for large numbers of CVs and higher-order effect calculations.

\begin{algorithm}[!ht]
  \DontPrintSemicolon
  \SetKw{KwSample}{sample\!}\SetKw{KwQuantile}{quantile\!}%
  \SetKwInOut{Input}{input}\SetKwInOut{Output}{output}%
  \SetKw{KwOutput}{output\!}%
  \SetKwComment{Comment}{$\triangleright$\ }{}%

  \Input{$\mathcal{M}$ \Comment*{Surrogate model
      \cref{eq:surrogate-model}}}%
  \Input{$N$ \Comment*{Number of replications}}%
  \Input{$M$ \Comment*{Number of observations per replication}}%
  \Input{$\delta$\quad
    $(\delta_l \defeq \delta/2, \delta_u\defeq 1-\delta_l)$
    \Comment*{Equal tail percentile level}} \BlankLine %
  \Output{$(\widehat{r}_1, \dots, \widehat{r}_p)$ \Comment*{Ranks
      \cref{eq:rank-est-1st}}}%
  \Output{$(\widehat{r}_1[\delta_l], \widehat{r}_1[\delta_u])\,,
    \dots\,, (\widehat{r}_p[\delta_l], \widehat{r}_p[\delta_u])$
    \Comment*{Percentile confidence intervals
      \cref{eq:rank-est-1st-ci}}}%
  \BlankLine %
   \Begin{%
    \emph{Compute $N$ replications of MISI first-order effect ranks}\;%
    \For{$n \in \{ 1, \dots, N\}$}{%
      \emph{Generate $M$ input-output samples}\;%
      \For{$m \in \{1, \dots, M\}$}{%
        \KwSample $(X_1^{(m)}, \dots, X^{(m)}_p)$\;
        $Y^{(m)} \leftarrow \mathcal{M}(X_1^{(m)}, \dots,
        X^{(m)}_p)$\;}%
       \emph{Calculate one replication of each MISI}\;
      \For{$j \in \{ 1, \dots, p\}$}{%
        $G(x,y) \leftarrow \log \bigl[ \widehat{f}_{X_j,Y}(x, y) \bigr]
      - \log \bigl[\widehat{f}_{X_j}(x) \widehat{f}_Y(y)\bigr] $ \;%

      $\widehat{s}^{\,(n)}_j \leftarrow \frac{1}{M} \sum_{1 \leq m \leq
        M} G(X_j^{(m)}, Y^{(m)})$ \Comment*{$n$th replication MISI
        \cref{eq:nth-rep-misi}}}%
    \emph{Calculate one replication of each rank}\;
    \For{$j \in \{1, \dots, p\}$}{%
      $\widehat{r}^{\,(n)}_{j} \leftarrow p - \#\{ \widehat{s}^{\,(n)}_{i}
      < \widehat{s}^{\,(n)}_j\,, \; i=1,\dots, p \}$ \Comment*{$n$th
        replication rank \cref{eq:rank}}%
    }%
  }%
  \BlankLine %
  \emph{Compute rank statistics from replications}\;%
  \For{$j \in \{ 1, \dots, p\}$}{%
    $\widehat{r}_j \leftarrow \frac{1}{N} \sum_{1\leq n \leq N}
    \widehat{r}^{\,(n)}_j$\;%
    $\widehat{r}_j[\delta_l] \leftarrow$ \KwQuantile
    ($\{\widehat{r}_j^{\,(1)}, \dots, \widehat{r}_j^{\,(N)}\},
    \delta_l$)\; $\widehat{r}_j[\delta_u] \leftarrow$ \KwQuantile
    ($\{\widehat{r}_j^{\,(1)}, \dots, \widehat{r}_j^{\,(N)}\}, \delta_u$)
  }%
  \Return{$(\widehat{r}_1, \dots, \widehat{r}_p)\,,
    (\widehat{r}_1[\delta_l], \widehat{r}_1[\delta_u])\,, \dots\,,
    (\widehat{r}_p[\delta_l], \widehat{r}_p[\delta_u])$
    \Comment*{cf.~\cref{alg:rank-cis} \KwOutput}} }%
\caption{Rank first-order MISIs with percentile confidence intervals}%
\label{alg:rank-dist-method}
\end{algorithm}

\section{Interrogating Black-Box Surrogates using MISIs}%
\label{sec:ginns}%

As highlighted in the introduction, our MI-based approach to GSA is applicable to any black-box surrogate model. To illustrate the ability of MI-based GSA to deal with correlated CVs, for which variance-based GSA approaches are of limited value, we combine it with a GINN, a domain-aware DNN surrogate introduced in~\cite{HallTaverniersEtAl:2020aa} to overcome computational bottlenecks in complex multiscale and multiphysics systems. \ADD{We consider two applications of interest in energy storage: the Langmuir adsorption model and a multiscale formulation of the dynamics of electrical double-layer supercapacitors with nanoporous electrodes.}

\subsection{\ADD{Langmuir adsorption model}}
\begin{figure}[!ht]
  \centering%
\includegraphics[width=0.9\textwidth]{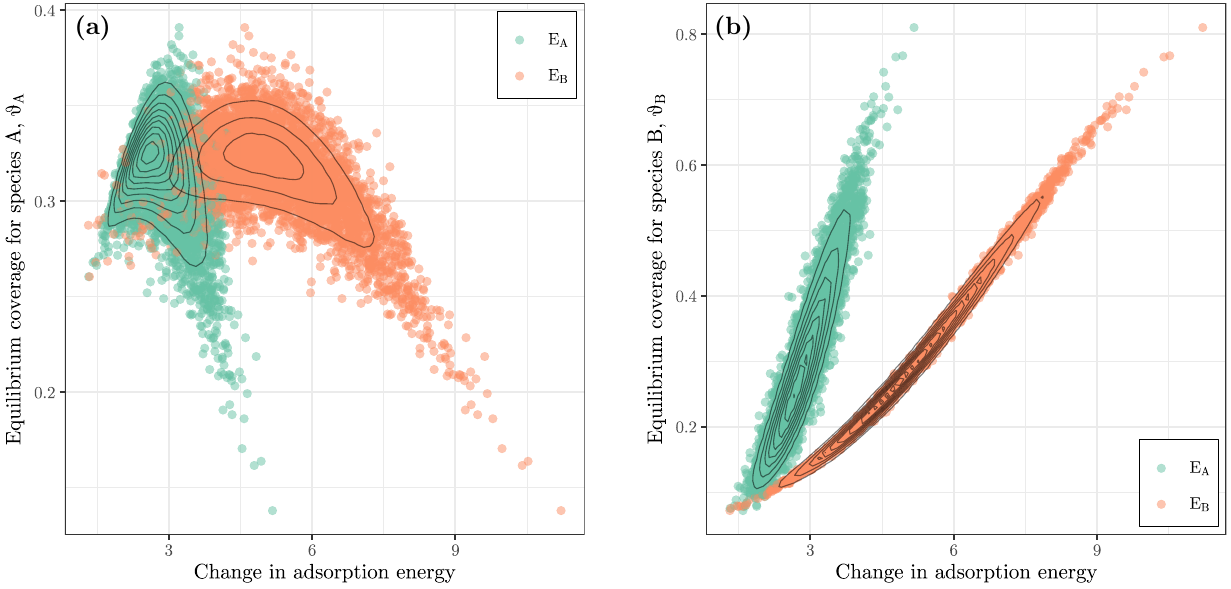}
\caption{\ADD{Response surfaces for the QoIs (a) $\vartheta_A$ and (b) $\vartheta_B$  (equilibrium coverages \cref{eq:langmuir-qoi}) based on inputs $E_A$ and $E_B$ (changes in adsorption energy \cref{eq:langmuir-inputs}) simulated using a physics-based model, i.e., the simplified statistical mechanics approximation \cref{eq:langmuir-simple} and the steady-state solution \cref{eq:langmuir-ss} for the Langmuir model \cref{eq:langmuir}. The equillibrium coverages are a nonlinear function of the changes in adsorption energy arising from an Arrhenius law \cref{eq:langmuir-arrhenius} and the calculation of the Gibbs free energy of adsorption \cref{eq:langmuir-gibbs}.}} \label{fig:langmuir-dat}
\end{figure}

\ADD{
To highlight the application of the MISI rankings, we first consider a simplified Langmuir adsorption model. This setting allows us to compare GINN-based rankings to ground truth rankings computed from the physics-based model. The Langmuir biomolecular adsorption model is widely used to describe competitive dissociative adsorption of two species on a catalyst surface \cite{DavisDavis:2003cr}, e.g., hydrogen oxidation in fuel cells \cite{GuXuYan:2014ox,LiuEtAl:2015ox,LeeMukerjeeEtAl:1999ox,NagasawaHanamura:2015ox}. The coverage dynamics $\vartheta_A$ and $\vartheta_B$ of species $A$ and $B$, respectively, are found by long-time integration of the system of nonlinear ODEs,
\begin{subequations}
    \label{eq:langmuir}
    \begin{align}
    &\frac{\dd \vartheta_A}{\dd t} = k^{\mathrm{ads}}_A P_A (1 - \vartheta_A - \vartheta_B)^2 - k^{\mathrm{des}}_A \vartheta_A^2\,, \quad &\vartheta_A(0) = \vartheta_A^0\,,\\
    &\frac{\dd \vartheta_B}{\dd t} = k^{\mathrm{ads}}_B P_B (1 - \vartheta_A - \vartheta_B)^2 - k^{\mathrm{des}}_B \vartheta_B^2\,, \quad &\vartheta_B(0) = \vartheta_B^0\,,
    \end{align}
\end{subequations}
where $P$ denotes partial pressure, $k^{\mathrm{ads}}$ denotes the adsorption rate constant, and $k^{\mathrm{des}}$ denotes the desorption rate constant. The system \cref{eq:langmuir} has a steady-state solution 
\begin{subequations}
    \label{eq:langmuir-ss}
    \begin{align}
    &\vartheta_A = \frac{(K_A P_A)^{1/2}}{1 + (K_A P_A)^{1/2} + (K_B P_B)^{1/2}}\,,\\
    &\vartheta_B = \frac{(K_B P_B)^{1/2}}{1 + (K_A P_A)^{1/2} + (K_B P_B)^{1/2}} \,,
    \end{align}
\end{subequations}
where $K = k^{\mathrm{ads}} / k^{\mathrm{des}}$ denotes an equilibrium constant. Our goal is to quantify the uncertainty in predicted coverages at equilibrium as functions of $E_A$ and $E_B$, the changes in adsorption energy of each species, 
\begin{equation}
    \label{eq:langmuir-qoi}
    \vartheta_A = \vartheta_A (E_A, E_B) \quad \text{and} \quad \vartheta_B = \vartheta_B (E_A, E_B)\,.
\end{equation}
These QoIs are highly nonlinear functions of $E_A$ and $E_B$; the equilibrium constants are expressed by the Arrhenius law,
\begin{subequations}
    \label{eq:langmuir-arrhenius}
    \begin{align}
    &K_A = \exp \left( - \frac{G_A}{k_B T} \right) (P_A + P_B)^{-1}\,,\\
    &K_B = \exp \left( - \frac{G_B}{k_B T} \right) (P_A + P_B)^{-1}\,,
    \end{align}
\end{subequations}
where $k_B$ is the Boltzmann constant; $T$ is the temperature; and the Gibbs free energy of adsorption,
\begin{align}
    \label{eq:langmuir-gibbs}
    G_A \propto -2E_A \quad\text{and}\quad
    G_B \propto -2E_B,
\end{align}
are, according to density functional theory, given by changes in adsorption energy plus additional terms that are not a function of the catalyst surface \cite{FengEtAl:2018uq}. The Gibbs free energy of adsorption is measured ``experimentally'' for a variety of metal catalyst surfaces via density functional theory, that is, quantum computations for actual metals. In \cite{FengEtAl:2018uq}, where competitive dissociative adsorption of $\mathrm{H_2}$ and $\mathrm{O_2}$ is considered, the changes in adsorption energy are observed to be correlated and are described by a linear model. 

We generate synthetic data, given in \cref{fig:langmuir-dat}, according to the following program. We assume the that $E_A$ is drawn from a Gamma distribution (with shape and scale)
\begin{subequations}
\label{eq:langmuir-inputs}
\begin{equation}
    E_A \sim \mathrm{Gamma}(\alpha = 33, \beta = 0.0870)\,.
\end{equation}
We then sample correlated $E_B$ according to the model
\begin{equation}
    E_B = -2 + 2.5 E_A + \epsilon\,,
\end{equation}
\end{subequations}
where each $\epsilon \sim \mathcal{N}(0,\sigma^2)$ is a mean zero noise. Assuming unit partial pressures, we consider the simplified statistical mechanics approximation
\begin{subequations}
\label{eq:langmuir-simple}
\begin{align}
    &G_A = -2E_A + 5\,, & G_B = -2E_B + 10\,,\\
    &K_A = \exp(-G_A/2)\,, & K_B = \exp(-G_B/2)\,.
\end{align}
\end{subequations}
Taking advantage of the closed form equilibrium solution \cref{eq:langmuir-ss}, we generate a corpus of $N_\mathrm{sam} = \num{1.25e6}$ input-output data $(X_{E_A}, X_{E_B}, Y_{\vartheta_A}, Y_{\vartheta_B})$  that allows us to construct ``ground truth'' rankings. As the input layer contains correlations, we consider a GINN \cite{HallTaverniersEtAl:2020aa} surrogate and learn two models: a fully resolved GINN that is trained and tested on the full corpus and a sparse GINN that is trained on a subset of $\num{1.25e4}$ data points. For simplicity of presentation we omit the full details on training and testing (see e.g.~\cref{sec:ginn-construction}).\footnote{We consider an architecture comprised of two hidden layers, each with 50 neurons and ReLU (Rectified Linear Unit) activations, and a linear output layer. In both cases, the training is set $80\%$ of the available data with remaining $20\%$ reserved for testing and the training and testing tolerance is $O(\num{1e-5})$.} 
}

\begin{figure}[!ht]
  \centering%
\includegraphics[width=0.9\textwidth]{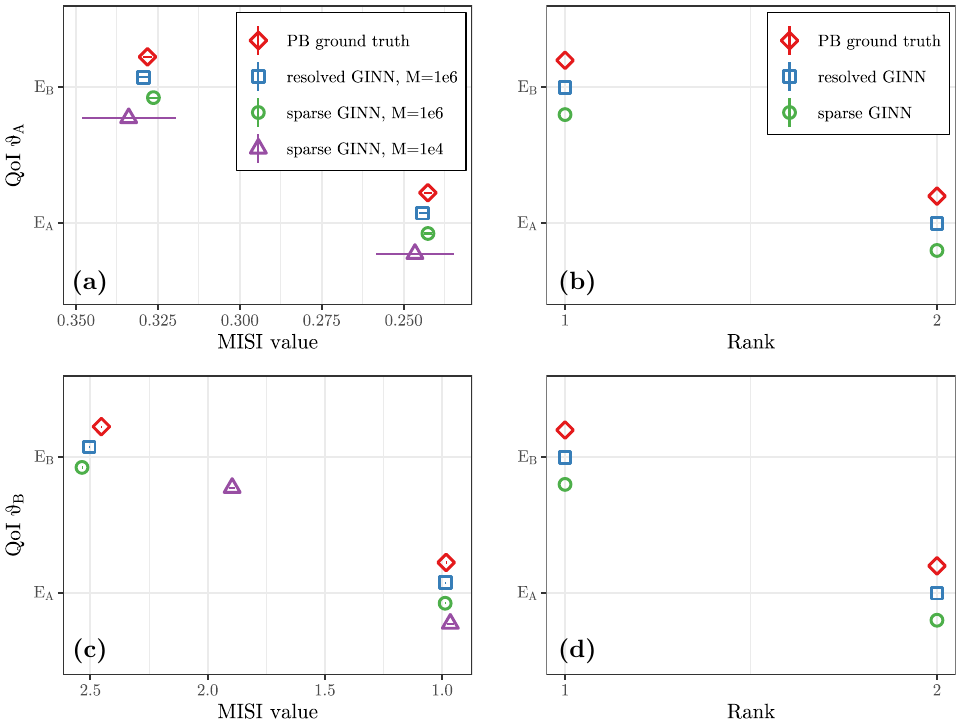}
\caption{\ADD{Plug-in Monte Carlo estimators of the first-order MISIs~\cref{eq:misi-est} for both a fully resolved GINN surrogate (learned from all available data) and a sparse GINN surrogate (learned from a restricted data set) are compared to a physics-based ground truth. For \cref{alg:rank-cis} (left column), we observe that the large-sample estimates based on $M=\num{1e6}$ samples are largely consistent. For \cref{alg:rank-dist-method} (right column), we observe that surrogate-driven mutual information-based GSA provides useful and consistent rankings. Note in the (b), (c), and (d) that the $95\%$ confidence intervals are vanishingly small.}} \label{fig:langmuir-rank}
\end{figure}

\ADD{
In \cref{fig:langmuir-rank}, we demonstrate the first-order MISI obtained using \cref{alg:rank-cis} and associated ranks calculated according to \cref{alg:rank-dist-method}, both with $95\%$ confidence intervals. In panels (a) and (c), estimated MISI \cref{eq:misi-est} based on $M=\num{1e6}$ samples from the physics-based model form a ``ground truth'' which are compared to the surrogate indices estimated using $M=\num{1e6}$ predictions from both the resolved and sparse GINN surrogate models. Although there is some bias in the learned surrogate models, the estimates based on large numbers of predictions are largely consistent with the ground truth rankings. In contrast, the estimate based on limited samples from the sparse GINN surrogate has larger confidence intervals (Figs.~\ref{fig:langmuir-rank}a and~\ref{fig:langmuir-rank}c), and the bias in sparse GINN surrogate severely underestimates the MISI value $I(E_B; \vartheta_B)$ (\cref{fig:langmuir-rank}c). All the estimated rankings resulting from \cref{alg:rank-dist-method} are consistent and have vanishingly small confidence intervals (Figs.~\ref{fig:langmuir-rank}b and~\ref{fig:langmuir-rank}d). The ground truth ranking and the resolved GINN estimates were computed with $N=\num{1e2}$ replications of $M=\num{1e3}$ samples and the sparse GINN model (trained on restricted data) was computed with $N = \num{10}$ replications of $M = \num{1e2}$ samples. Taken together, this experiment demonstrates that, although bias or error in the trained GINN can impact predictions, the surrogate-driven mutual information-based GSA provides useful rankings. Although evaluating the information flow between the physics-based model and its surrogate is beyond the scope of the present work, we point to an analysis~\cite{TishbyZaslavsky:2015bn} of the information theoretic limits of DNNs, assuming the inputs have a known model and are conditionally independent. 
}

\subsection{Multiscale supercapacitor dynamics}
\label{sec:testbed}

We consider an electrical double-layer capacitor \cite{Soffer:1972sc}, whose electrodes are made of a conductive hierarchical nanoporous carbon structure \cite{Narayanan:2016sc}. Electrolyte (an ionized fluid) fills the nanopores and contributes to the formation of the electrical double layer at the electrolyte-electrode interface (see, e.g., Fig.~A8 in~\cite{HallTaverniersEtAl:2020aa}).  Identification of an optimal pore structure of the carbon electrodes holds the promise of manufacturing supercapacitors which boast high power and high energy density~\cite{Nomura:2019sc,LiEtAl:2020sc}. This and other advancements, such as lower self-discharge electrolytes \cite{Wang:2019sc} for enhanced long-term energy storage, position electrical double-layer capacitors as a viable replacement of Li-ion batteries in electric vehicles or personal electronic devices. Attractive features of electrical double-layer capacitors are their shorter charging times, longer service life, and reduced reliance on hazardous materials~\cite{Beguin:2013sc}.

Two macroscopic QoIs affect electrical double-layer capacitor performance: effective electrolyte conductivity $\keff$ and transference number  $\tp$ (fraction of the current carried by the cations), such that $\bm Y \defeq \{ Y_{\keff}, Y_{\tp} \}$. These QoIs are influenced by seven tunable CVs: the electrode surface (fluid-solid interface) potential $\phiG$, initial ion concentration $\cin$, temperature $T$, porosity $\omega$, (half) pore throat size $\lpor$, solid radius $r$, and Debye length $\lambda_D$, such that $\bm{X} \defeq \{X_\phiG, X_{\cin}, X_T, X_\omega, X_{\lpor}, X_r, X_{\lambda_D}\}$. A physics-based model $\mathcal M$, derived in~\cite{ZhangTartakovsky:2017np} via homogenization, relates the inputs $\bm X$ to the outputs $\bm Y$. This model involves closure variables (second-order tensors) $\bchi_{\pm}$ and electrical double-layer potential $\phiEDL$, whose determination is expensive and constitutes computational bottlenecks  $\bm{Z} \defeq \{Z_{\bchi_{\pm}}, Z_{\phiEDL}\}$.
Optimal design of the nanoporous electrodes in electrical double-layer capacitors involves the tuning of the CVs $\bm X$ to elicit changes in the QoIs $\bm Y$. %

\begin{figure}[!ht] 
\centering 
\includegraphics[width=0.6\textwidth]{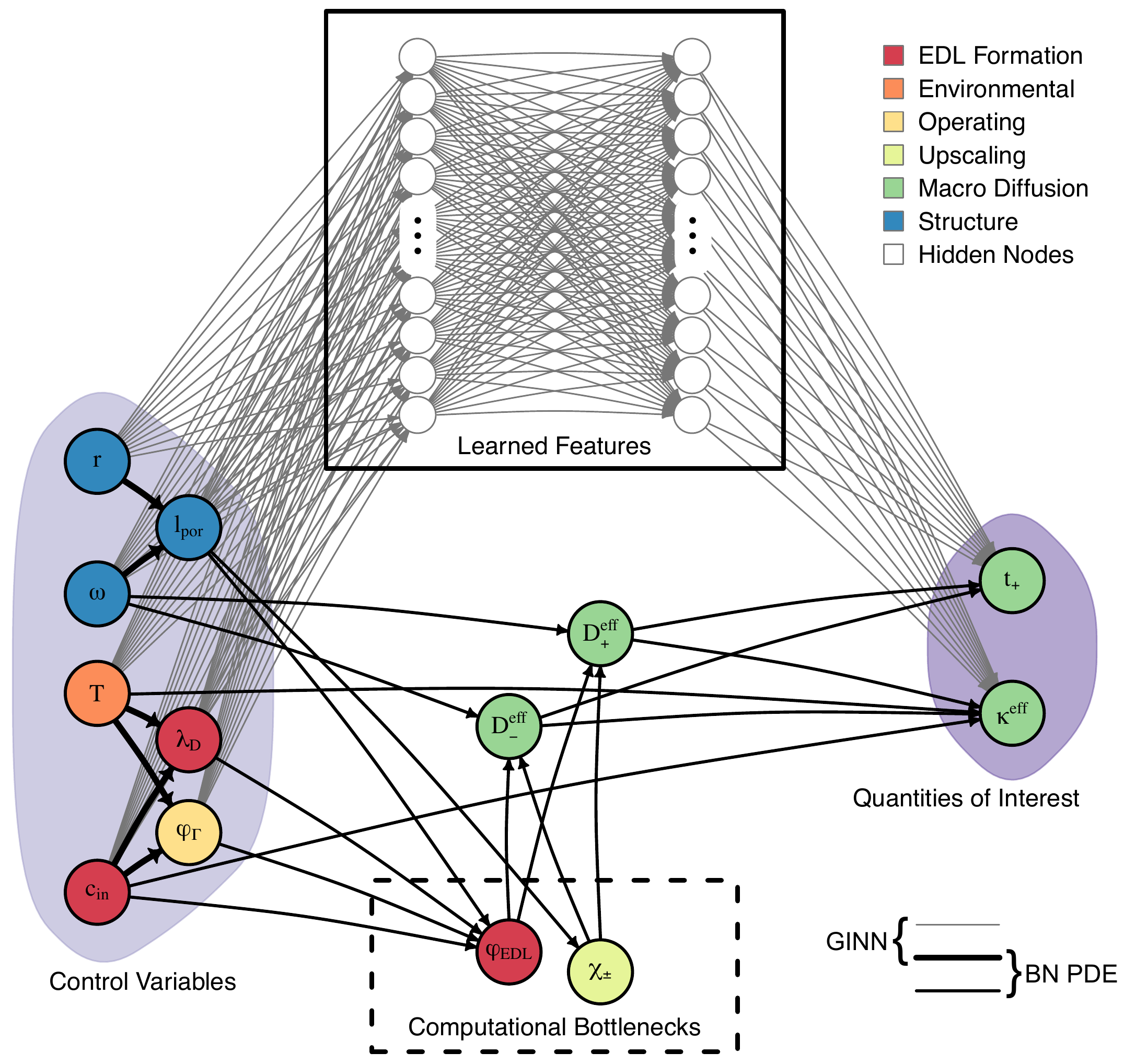}
  \caption{Visualization of the BN PDE (lower route) and GINN surrogate (upper route) for a multiscale model of EDL supercapacitor dynamics. The BN encodes conditional relationships between the model variables (both inter- and intrascale) and systematically includes domain knowledge into the physics-based model, ensuring the resulting BN PDE makes physically sound predictions. The GINN takes identical inputs $\bm X$ (i.e., structured priors on CVs) to those of the BN PDE, but overcomes the latter's computational bottlenecks $\bm{Z}$ (dashed box) by replacing them with learned features (solid box) in a DNN to predict the QoIs $\bm Y$. The nodes in the hidden layers of the GINN make it a black box.}
  \label{fig:BN_NN}
\end{figure}

The complex nonlinear and multiscale relationship between $\bm X$ and $\bm Y$ makes this a challenging engineering design problem and allows us to highlight the features (i)--(iii) of the MI-based  GSA. The joint PDF of the random CVs $\bm X$ systematically quantifies uncertainties and errors arising in the physics-based representation. This key quantity for decision support is captured by a Bayesian Network (BN)~\cite{HallTaverniersEtAl:2020aa,UmHallEtAl:2019bn}, which encodes both physical relationships and available domain knowledge (\cref{fig:BN_NN}). The resulting probabilistic physics-based model $\mathcal M$, referred to as a BN PDE, propagates the joint PDF of $\bm X$, i.e., a structured prior, via $\bm{Z}$ to $\bm{Y}$ following the conditional relationships in the BN.
As in \cite{HallTaverniersEtAl:2020aa}, we assume both the CVs $X_{T}$, $X_{\cin}$, $X_{r}$, and $X_{\omega}$ to be independent and their prior PDFs to be uniform on an interval of $\pm 35 \%$ (for $X_{T}$ and $X_{\cin}$) or $\pm 25 \%$ (for $X_{r}$ and $X_{\omega}$) around their respective baseline values (see \cref{tab:independent-params} reproduced from \cite{HallTaverniersEtAl:2020aa}),
\begin{equation}
  \label{eq:inputs-independent} X_{i} \given \theta_{i} \sim \unif([\theta_{i}^{\min}, \theta_{i}^{\max}]),
  \qquad i = T, \cin, r, \omega;
\end{equation} 
where the hyperparameters $\theta_i = \{\theta_{i}^{\min}, \theta_{i}^{\max}\}$ represent the left and right endpoints of the support intervals. The remaining CVs, $X_{\lambda_D}$, $X_{\phiG}$ and $X_{\lpor}$,  are related conditionally (\cref{fig:BN_NN}) to these independent inputs through the physical relations,
\begin{subequations}
  \label{eq:correlated_CVs}
  \begin{align}
    \label{eq:lambda_D} \lambda_D &= \sqrt{\frac{RT\mathcal{E}}{2F^2 z^2 \nu \cin}} \quad [\si{\nano\meter}],\\
    \label{eq:phiG} \phiG &= \frac{V}{2} - \varphi_\text{ecm} - \frac{1}{C_\text{H}}\sqrt{4\mathcal{E}
        RTz^2\cin}\sqrt{\cosh\left(\frac{\mathrm{e}\phiG}{k_{\text{B}}T}\right)
        -
        \cosh\left(\frac{\mathrm{e}\varphi_\text{min}}{k_{\text{B}}T}\right)} \quad [\si{\volt}],\\
    \label{eq:lpor} \lpor &= -r + 0.5\sqrt{4r^2 + 4r^2 \left[\frac{\pi}{4\cdot (1 - \omega)} - 1\right]} \quad [\si{\nano\meter}]\,.
  \end{align}
\end{subequations}
\ADD{Here, $F = 96485$~\si{\coulomb\per\mol} is the Faraday constant, $R$~[\si{\joule\per\mol\per\kelvin}] is the gas constant, $k_{\text{B}}$~[\si{\joule\per\kelvin}] is the Boltzmann constant,  $T$~[\si{\kelvin}] is the temperature, $\mathcal{E}$~[\si{\farad\per\meter}] is the absolute permittivity of the solvent, $z$~[-] is the ion charge (valence), $\nu$~[-] is the dissociation constant, $V$ is the external voltage, $\mathrm{e}$~[\si{\coulomb}] is the elementary charge, $\varphi_\text{ecm}$ is the electrocapillary maximum, $\cin$~[\si{\mole\per\liter}] is the initial ion concentration, and $\varphi_\text{min}$ is the midplane potential  (see \cite{HallTaverniersEtAl:2020aa} for further details).}
The PDFs of these dependent CVs are estimated by sampling the uniform distributions \cref{eq:inputs-independent} and computing a corresponding observation via \cref{eq:correlated_CVs}. Hence, the physics of the problem induces the correlations between the CVs represented by the conditional relationships in~\cref{fig:BN_NN}.

\begin{table}[htbp] 
  \centering
  \caption{Statistics of the uniform PDFs of the independent CVs in~\cref{eq:inputs-independent} (from
    \cite{HallTaverniersEtAl:2020aa}).}
  \label{tab:independent-params}
  \begin{tabular}{cSSScc}
    \toprule%
    \multicolumn{1}{l}{Variable label} & $\theta^{\min}$ & $\theta^{\max}$ & {Mean/Baseline} & {Variation} & {Units}\\%
    \midrule%
    $T$  & 208 & 432 & 320 & $\pm 35\%$ & \si{\kelvin} \\%
    $\cin$  & 0.52 & 1.08 &  0.80 & $\pm 35\%$ & \si{\mol\per\liter} \\%
    $r$  & 1.05 & 1.75 &  1.40 & $\pm 25\%$ & \si{\nano\meter} \\
    $\omega$  & 0.5025 & 0.8375 & 0.6700 & $\pm 25\%$ & - \\
    \bottomrule
  \end{tabular}
\end{table}

\subsection{GINNs: surrogate models for multiscale physics}
\label{sec:GINN_multiscale}

GINNs \cite{HallTaverniersEtAl:2020aa} are domain-aware surrogates for a broad range of complex physics-based models. In the context of electrical double-layer capacitors, a GINN can be used to accelerate the propagation of uncertainty from structured priors on CVs $\bm{X}$ to distributions of QoIs $\bm{Y}$ by replacing the computational bottlenecks $\bm{Z}$ in the BN PDE (the dashed boxed nodes of the BN in \cref{fig:BN_NN}) with the GINN's hidden layers. %
In so doing, it alleviates the cost of computing the QoIs $\bm Y$, which includes bypassing the need to compute the effective diffusion coefficients of the cations ($\Deffp$) and anions ($\Deffm$) according to
\begin{subequations}
\begin{align}
  \label{eq:kappa-eff} 
  \keff \defeq \; & \nu z^2 \frac{F^2 \cin}{RT} (\Deffp + \Deffm)\ \quad [\si{\milli\siemens\per\cm}]\,,\\
  \label{eq:transference}
  \tp \defeq \; & \frac{\Deffp}{\Deffp + \Deffm}\ \quad  [-]\,.
\end{align} 
\end{subequations}

\subsubsection{GINN construction} \label{sec:ginn-construction}

The workflow for building the GINN surrogate for the supercapacitor dynamics example is summarized as follows. Details, including the procedures for training and testing the GINN using the BN PDE, can be found in \cite{HallTaverniersEtAl:2020aa}.
\begin{enumerate}
\item \textbf{Data generation (BN PDE):} Generate $\Nsam=\num{4e3}$ input-output samples by drawing the inputs from the structured priors on $\bm{X}$ and computing the corresponding responses $\bm{Y}$ with the BN PDE, and select $\Ntrain=0.75\Nsam$ training samples and $\Ntest=0.25\Nsam$ test samples from this data set.
\item \textbf{Training:} Using the $\Ntrain$ input-output pairs and \texttt{TensorFlow 2}, train with 100 epochs a fully connected NN comprising: %
  \begin{compactenum}[a)]
  \item an input layer consisting of the seven CVs $\bm X$,
  \item two hidden layers each consisting of 100 neurons, 
  \item an output layer consisting of the two QoIs $\bm Y$,
  \item application of the ReLU (Rectified Linear Unit) activation function, and
  \item a given training error tolerance of $10^{-4}$.
  \end{compactenum} 
\item \textbf{Testing:} Test the trained GINN on the $\Ntest$ input-output pairs to analyze its generalization capability for unseen data for a given test error tolerance of $10^{-4}$.
\item \textbf{Prediction:} Sample $\Nsam^\text{pred}$ inputs from the structured priors on the CVs, and predict the corresponding responses with the trained GINN.
\end{enumerate} 

\subsubsection{Computational efficiency of the physics- and GINN-based models}

For complex numerical simulations, the cost of step 1 outweighs, by orders of magnitude, the combined cost of steps 2--4~\cite{HallTaverniersEtAl:2020aa}. The GSA results reported below require $\num{3e3}$ samples of the BN PDE (physics-based model) to train the GINN that satisfies both the training and test error tolerances.\footnote{While testing requires the generation of $\num{1e3}$ additional input-output samples with the BN PDE, it is not strictly required and hence not taken into account when comparing the rankings generated with the physics-based model and the GINN.} Since the generation of new samples with the GINN carries a negligible expense compared to the generation of training data with the BN PDE, the computational costs of the physics- and GINN-based GSAs are virtually identical when carrying out the former using $\num{3e3}$ samples; this allows us to investigate the performance of both approaches for a fixed computational budget. 

\subsection{GINN-based MISI rankings for supercapacitor dynamics}
\label{sec:rankings_testbed}

\begin{figure}[!ht]
  \centering%
\includegraphics[width=0.9\textwidth]{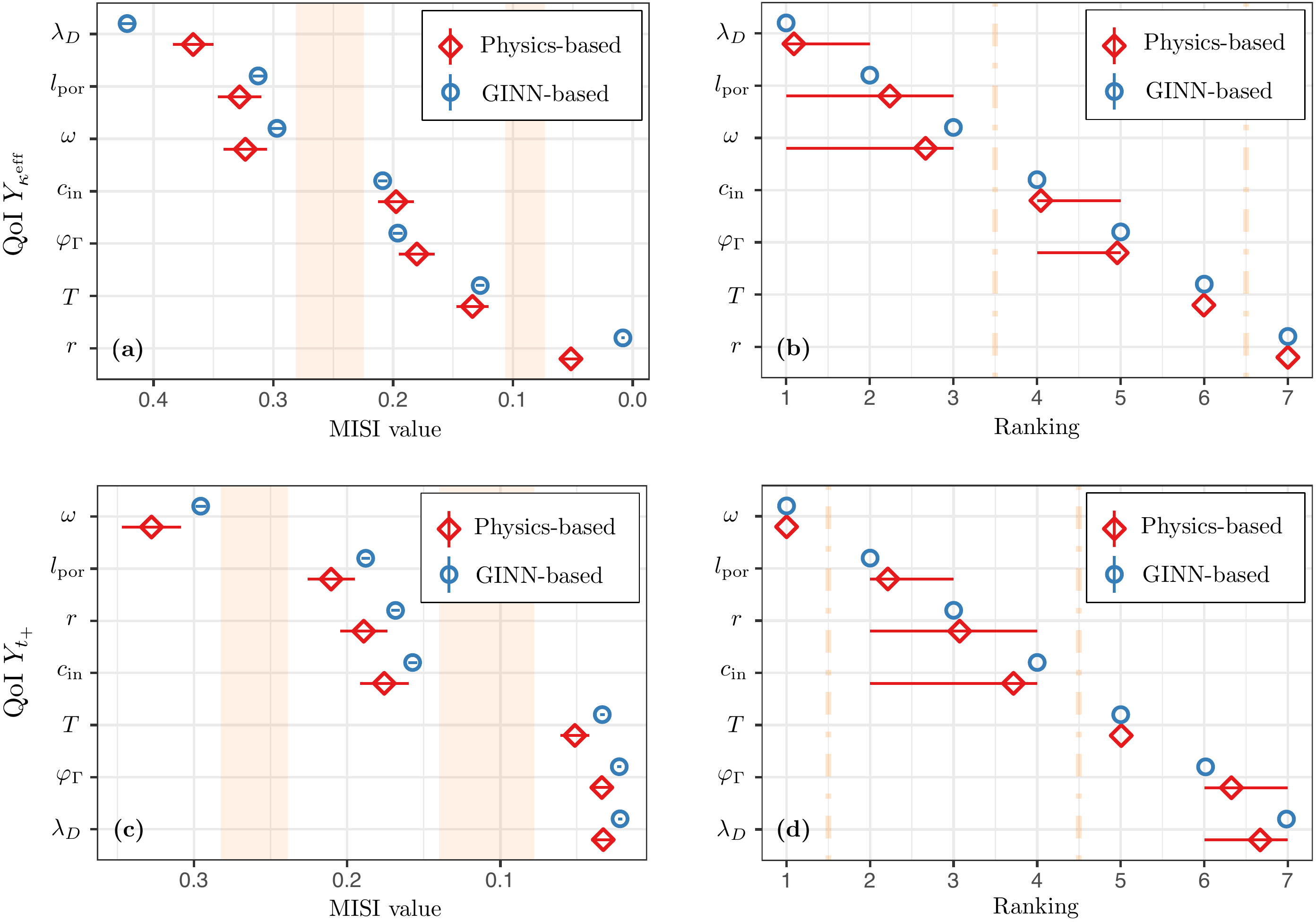}
\caption{Plug-in Monte Carlo estimators of the first-order MISIs in~\cref{eq:misi-est} indicate the most impactful CVs $\bm X$ for tuning the QoIs $Y_\keff$ (top row) and $Y_\tp$ (bottom row). For \cref{alg:rank-cis} (left column), the width of the confidence intervals~\cref{eq:rank-1st-ci-adjusted} is chosen to achieve a non-overlap significance level $\bar{\gamma} = 0.01$ pairwise on average. The highlighted gaps in (a) and (c) indicate clusters of CVs with similar relative importance. For \cref{alg:rank-dist-method} (right column), the GINN-based estimators for CV ranking with percentile confidence intervals \cref{eq:rank-est-1st-ci} are consistent with the rankings in (a) and (c). These ranks are resolved, which is not feasible through bootstrapping the $\num{3e3}$ samples from the physics-based model. In both algorithms, the GINN surrogate enables querying sufficiently large amounts of data to distinguish closely-ranked CVs with a high degree of confidence \ADD{(relative to bootstrapping)}. In (b) and (d) the dashed lines correspond to the gaps identified in (a) and (c), respectively, \ADD{and we observe that some of the confidence intervals related to the GINN-based estimates are vanishingly small}.} \label{fig:ranking-1st}
\end{figure}

Figure~\ref{fig:ranking-1st} exhibits the first-order MISI values for $Y_\keff$ and $Y_\tp$ (left column) and the corresponding ranks of the CVs $\bm X$ (right column), estimated respectively with \cref{alg:rank-cis} and \cref{alg:rank-dist-method}. All of these quantities are computed, alternatively, with the physics- and GINN-based models. The MISI values are equipped with the adjusted confidence intervals indicating a pairwise non-overlap significance $\bar{\gamma} = 0.01$ (on average, at level $\beta \approx 0.05$). The 95\% percentile confidence intervals for the CV ranks in \cref{eq:rank-est-1st-ci}, i.e., with $\delta=0.05$, are based on $N=\num{1e3}$ replications; samples for the estimator in each replication are either predicted using the GINN or are bootstrap resampled from a corpus of $\num{3e3}$ physics-based simulations. For both  \cref{alg:rank-cis} and \cref{alg:rank-dist-method}, the physics- and GINN-based estimators are largely consistent, which is to be expected since the GINN surrogate satisfies both a preset training and test error tolerance. The highlighted gaps between clusters of MISI values in~\cref{fig:ranking-1st}a,c indicate the groupings of various CVs $\bm X$ by their relative importance. In~\cref{fig:ranking-1st}b,d dashed lines correspond to these highlighted gaps; although the clarity of the rankings in \cref{fig:ranking-1st}b,d facilitates the automation of decisions in outer-loop tasks, the ranks themselves do not contain information about the relative importance of each parameter as in \cref{fig:ranking-1st}a,c.

For both QoIs, the MISI estimators obtained with the physics-based model lead to indeterminate rankings. For $Y_\keff$, the confidence intervals for $\{ X_\lpor , X_{\omega}\}$ and $\{X_{\cin}, X_{\phiG}\}$ overlap, and therefore the difference between their MISI values (i.e., their ranking) does not differ at an average significance level $\bar{\gamma} = 0.01$. Similarly, for $Y_\tp$ the rankings for $\{ X_\lpor, X_{r}, X_{\cin}\}$ are not resolved. In contrast, the differences between the corresponding estimators derived from $\num{5e4}$ (for $Y_\keff$) or $\num{1e5}$ (for $Y_\tp$) GINN-based predictions are pairwise significant at the level $\bar{\gamma} = 0.01$. Likewise, we observe that the ranks generated using $N = \num{1e3}$ replications with $M = \num{5e4}$ observations for $Y_\keff$ or $M = \num{1e5}$ observations for $Y_\tp$ predicted with the GINN are fully resolved (indeed, the 95\% percentile confidence intervals are vanishingly small on the plots). Moreover, these ranks are consistent with the resolved rankings of the MISI values for $Y_\keff$ and $Y_\tp$ deduced from the GINN-based estimators obtained with \cref{alg:rank-cis}. These findings demonstrate the benefit of a GINN, since resolving the rankings with the physics-based model is considerably more expensive given the high cost of generating additional response samples with the BN PDE.

Figure~\ref{fig:ranking-1st} also compares the direct distributional method of \cref{alg:rank-dist-method} to bootstrapped estimators with percentile confidence intervals computed using the physics-based model. Using $N = \num{1e3}$ bootstrap (i.e.,~resampling~\cite{Wasserman:2006np}) replications of $M = \num{3e3}$ observations ($M$ is constrained by the fixed computational budget) yields clusters of indeterminate ranks: the ranking of $\{X_{\lambda_D}, X_{\lpor}, X_{\omega} \}$ and $\{X_{\cin}, X_{\phiG}\}$ for $Y_\keff$, and the ranking of $\{X_{\lpor}, X_{r}, X_{\cin}\}$ and $\{X_{\phiG}, X_{\lambda_D}\}$ for $Y_{\tp}$, cannot be resolved at the $0.05$ level. In both cases, this is over $70\%$ of the CVs. Again, that ranking of the first-order effects of the CVs with a high degree of confidence \ADD{(relative to bootstrapping)} within a constrained computational budget critically depends on the availability of a GINN (or, more generally, a surrogate model) to cheaply generate additional response samples.

\begin{figure}[!ht]
  \centering%
   \includegraphics[width=0.8\textwidth]{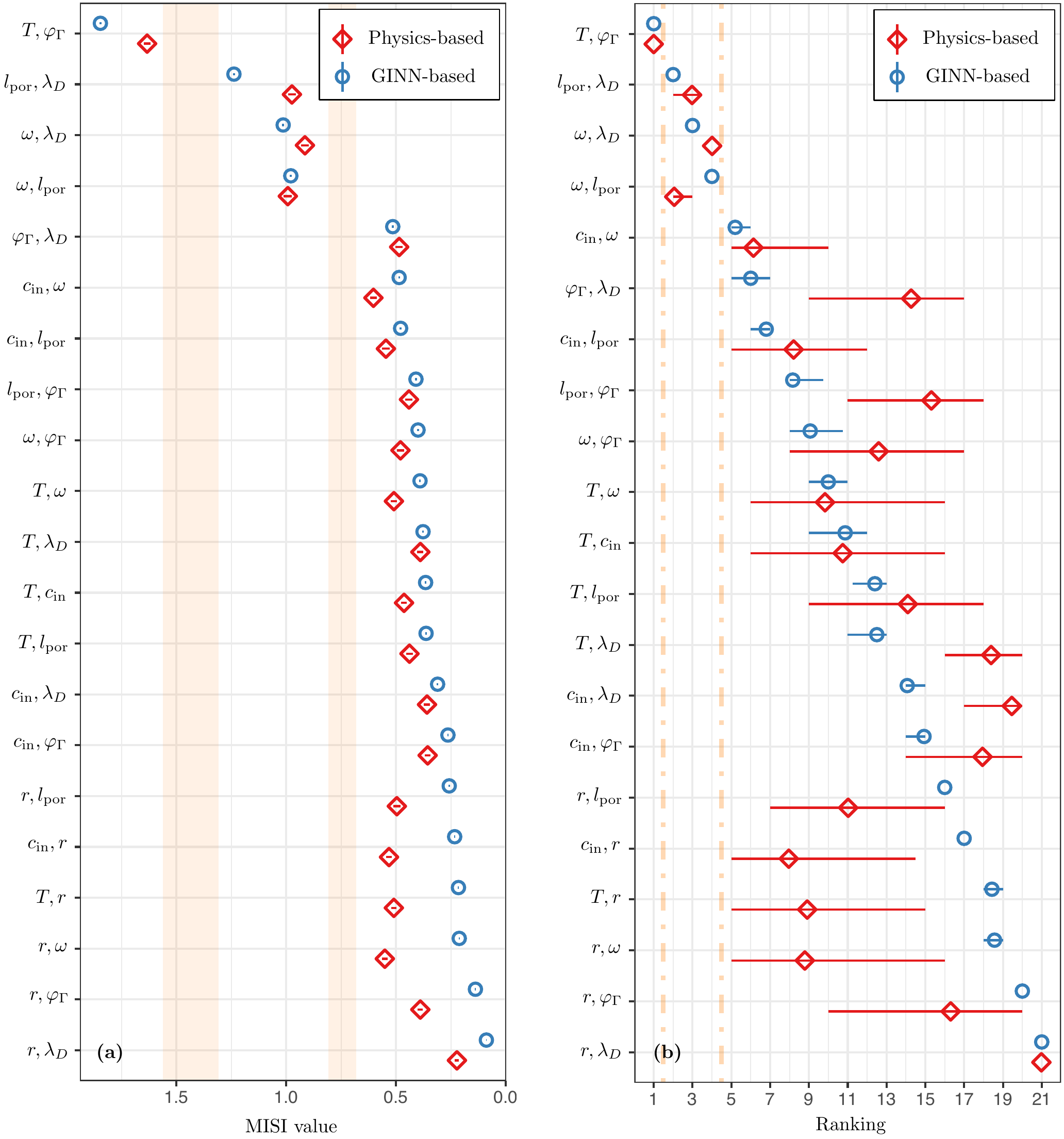}
   \caption{Plug-in Monte Carlo estimators of the second-order MISIs in~\cref{eq:misi-2nd-est} indicate the most impactful interactions between any two CVs in $\bm X$ for tuning the QoI $Y_\keff$. The GINN surrogate improves the resolution of the second-order MISI rankings \ADD{(relative to bootstrapping)} computed with (a) \cref{alg:rank-cis} or (b) \cref{alg:rank-dist-method}. Although the first-order effect of $X_T$ and $X_{\phiG}$ on $Y_\keff$ are not top-ranked (see \cref{fig:ranking-1st} (a) and (b)), we observe above that $(X_{T},X_{\phiG})$ has the most important second-order effect on $Y_\keff$.}
     \label{fig:ranking-2nd}
\end{figure}

Since the CVs $\bm X$ are correlated, it is natural to expect higher-order effects due to interactions between the CVs. Figure~\ref{fig:ranking-2nd} displays the second-order MISIs and their ranks estimated using \cref{alg:rank-cis,alg:rank-dist-method}, respectively, for the QoI $Y_{\keff}$. The pairwise comparison of the adjusted confidence intervals with non-overlap significance $\bar{\gamma} = 0.01$ reveals that approximately 80\% of the estimators obtained with $M=\num{3e3}$ physics-based observations are indistinguishable (\cref{fig:ranking-2nd}a). In contrast, the rankings deduced from the estimators obtained with $M=\num{5e4}$ GINN-based observations are fully resolved. The estimated second-order effect ranks based on $N = 30$ replications of $M = \num{1e5}$ GINN-based observations are nearly identical, emphasizing the robustness and consistency of \cref{alg:rank-cis,alg:rank-dist-method}. A large proportion of the MISI values is clustered and equally important. The GINN-based GSA resolves $\approx 40\%$ of the ranks compared to $\approx 15\%$ of the ranks distinguishable from $N = \num{1e3}$ bootstrap replications of $M = \num{3e3}$ physics-based observations. Here, the number of replications for the GINN-based ranks is chosen such that the total computational time of implementing \cref{alg:rank-dist-method} is similar to that of computing all the physics-based bootstrap replications, which is approximately the case when the product $N\cdot M$ is the same for both models.

To summarize the key findings from the numerical experiments presented in \cref{fig:ranking-1st,fig:ranking-2nd}: %
the GINN surrogate %
\ADD{enables the fast generation of useful distinguishable rankings}. These rankings are largely consistent with the budget-constrained predictions of the physics-based model. Hence, \cref{alg:rank-cis,alg:rank-dist-method} facilitate the deployment of GINN for the acceleration and future automation of outer-loop decision-support tasks.

\section{Design with Explainable Black-Box Surrogates }%
\label{sec:misi-validation}%

Like all deep neural networks, GINNs are black boxes that lack a clear functional relationship between inputs and outputs. MI-based GSA aids in interpreting and explaining their predictions, thereby enabling the use of black-box surrogates in simulation-based decision-making, including the closure of engineering design loops to facilitate rapid prototyping.

We validate the first-order MISI rankings discussed in \cref{sec:rankings_testbed}, and then use these rankings to explore subregions of the original parameter space that deliver high values of the effective electrolyte conductivity, $\keff$. Subsequent effect rankings within this parameter subspace suggest follow-up simulations or novel laboratory experiments, resulting in further refinements to the design of nanoporous electrodes for electrical double-layer capacitors.

\subsection{Validation of MISI rankings \ADD{for supercapacitor dynamics}}

\begin{figure}[htb]
  \centering%
    \includegraphics[width=0.8\textwidth]{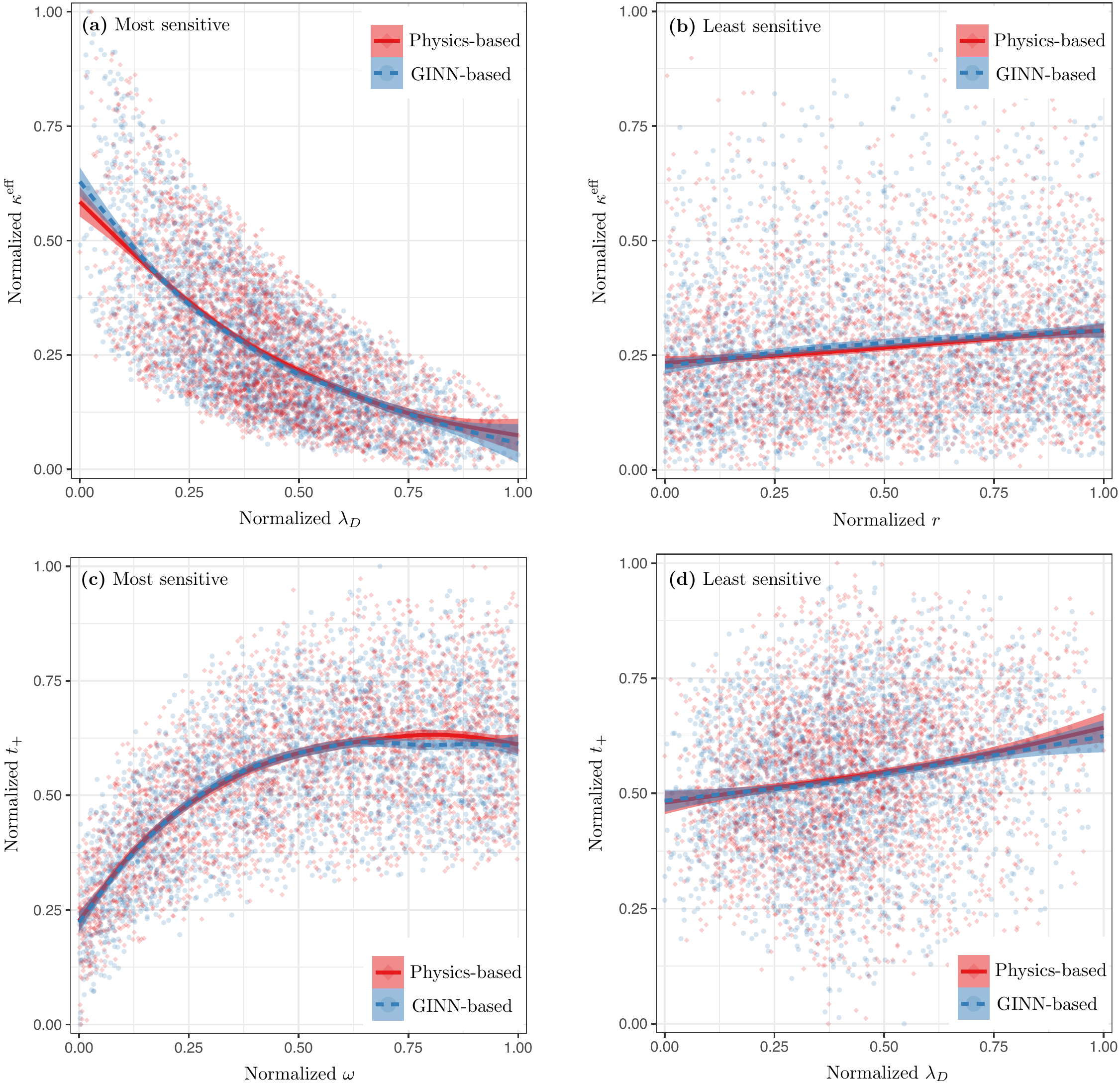}
  \caption{Response surfaces of the QoIs $\kappa^\text{eff}$ (top row) and $t_+$ (bottom row) with respect to the respective most (left column) and least (right column) sensitive parameters. The plots represent $\num{3e3}$ observations which are fitted with cubic regression splines. These results indicate nonlinear response surfaces for the most sensitive parameter directions, (a) and (c), in contrast to the random dispersion of observations for the least sensitive parameter directions, (b) and (d). This  validates the first-order effect rankings in \cref{fig:ranking-1st}.} \label{fig:validation}
\end{figure}

\Cref{fig:validation} shows normalized response surfaces, in the form of scatter plots and cubic regression splines based on $\num{1e3}$ observations, for the QoIs $\keff$ and $\tp$ along sensitive and insensitive parameter directions identified by the first-order MISI rankings in \cref{fig:ranking-1st}. The most sensitive parameter directions are $\lambda_D$ for $\keff$ and $\omega$ for $\tp$, and the least sensitive are $r$ for $\keff$ and $\lambda_D$ for $\tp$. The response surfaces for $\keff$ and $\tp$ (top and bottom rows in \cref{fig:validation}, respectively) demonstrate nonlinear relationships with respect to the most sensitive CV directions (Figs.~\ref{fig:validation}a,c). In contrast, the random scatter for the least sensitive directions (Figs.~\ref{fig:validation}b,d) suggests the lack of a clear relationship between these CVs and the QoIs. The quality and strength of these functional relationships validate the assigned rankings.

The MISI effect ranking and above validation step lend interpretability to the black-box predictions. This  enables the use of GINNs in design iterations by predicting new response samples in reduced parameter spaces that optimize certain QoIs. The next section illustrates this procedure.

\subsection{Design of multiscale systems under uncertainty}

\begin{figure}[phtb]
  \centering%
\includegraphics[width=0.8\textwidth]{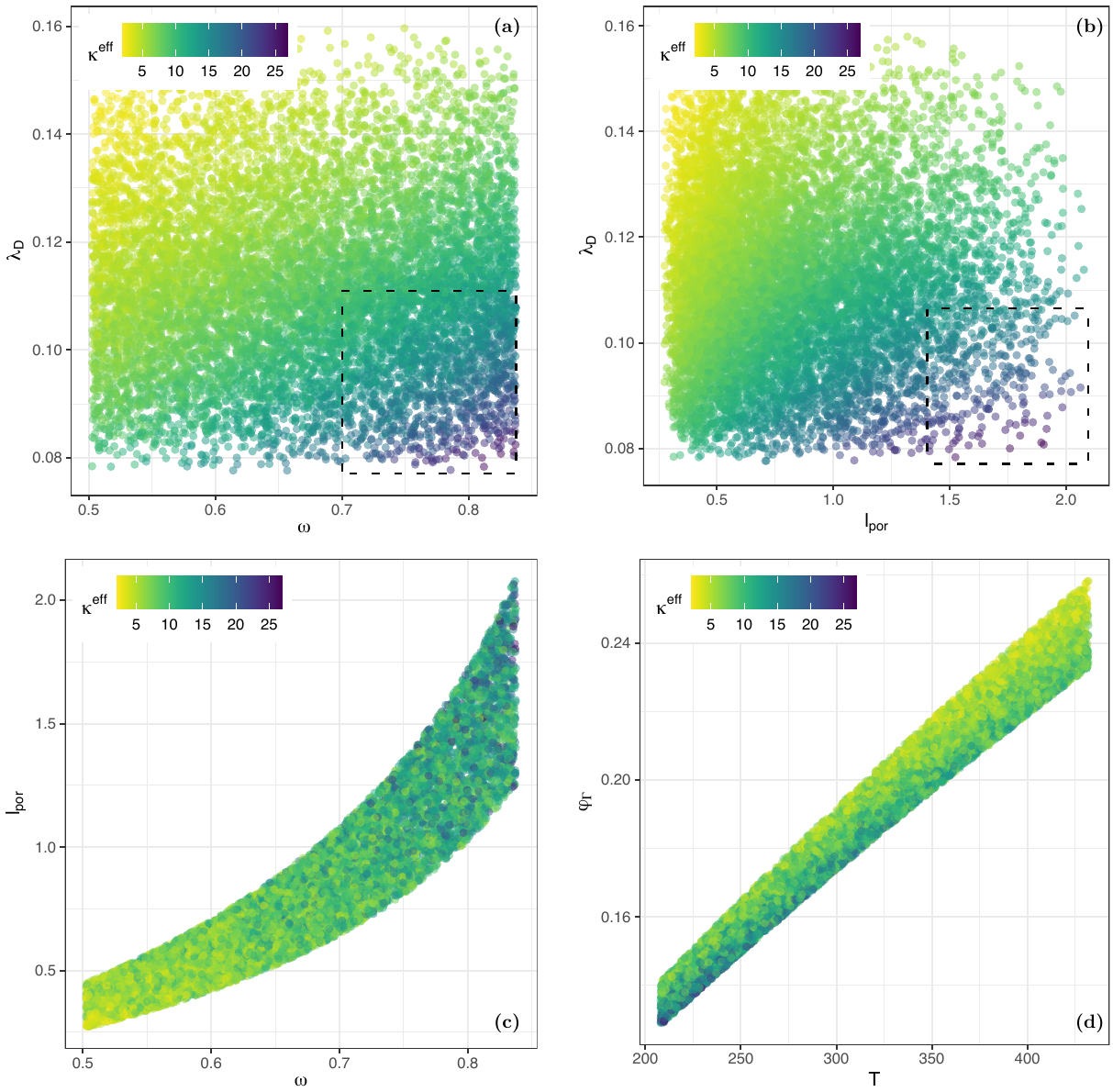}
  \caption{GINN-predicted response surfaces of the effective electrolyte conductivity $\keff$ based \num{3e3} observations identify reduced parameter ranges corresponding to a targeted response which can be explored to close design loops. The response surfaces are plotted over two-dimensional subspaces of the full input space; in (a), (b), and (c) the CVs correspond to the top-ranked first-order MISIs in \cref{fig:ranking-1st} and in (d) they correspond to the highest second-order MISI in \cref{fig:ranking-2nd}. In (a) and (b) the subspaces enclosed by dashed boxes correspond to high values of $\keff$ and are the focus of the further investigations in \cref{fig:close_design_loop_rank}. 
    }
  \label{fig:close_design_loop}
\end{figure}

The first- and second-order MISI rankings suggest that the CVs $\lambda_D$, $\lpor$, and $\omega$ have the largest individual contributions to changes in $\keff$ (\cref{fig:ranking-1st}), and the CV pairs $(T,\phiG)$, $(\lpor, \lambda_D)$, $(\omega, \lambda_D)$ and $(\omega, \lpor)$ (\cref{fig:ranking-2nd}) have the largest pairwise interaction effect. The GINN-generated response surfaces of $\keff$ for these CV pairs (\cref{fig:close_design_loop}) identify the parameter subspaces in which a targeted range of the QoI $\keff$, e.g., its maximal value, is likely to be achieved. Computation of new MISI rankings in these restricted subregions determines which CVs to retain in the next design cycle. 

\begin{figure}[phtb]
  \centering%
\includegraphics[width=0.8\textwidth]{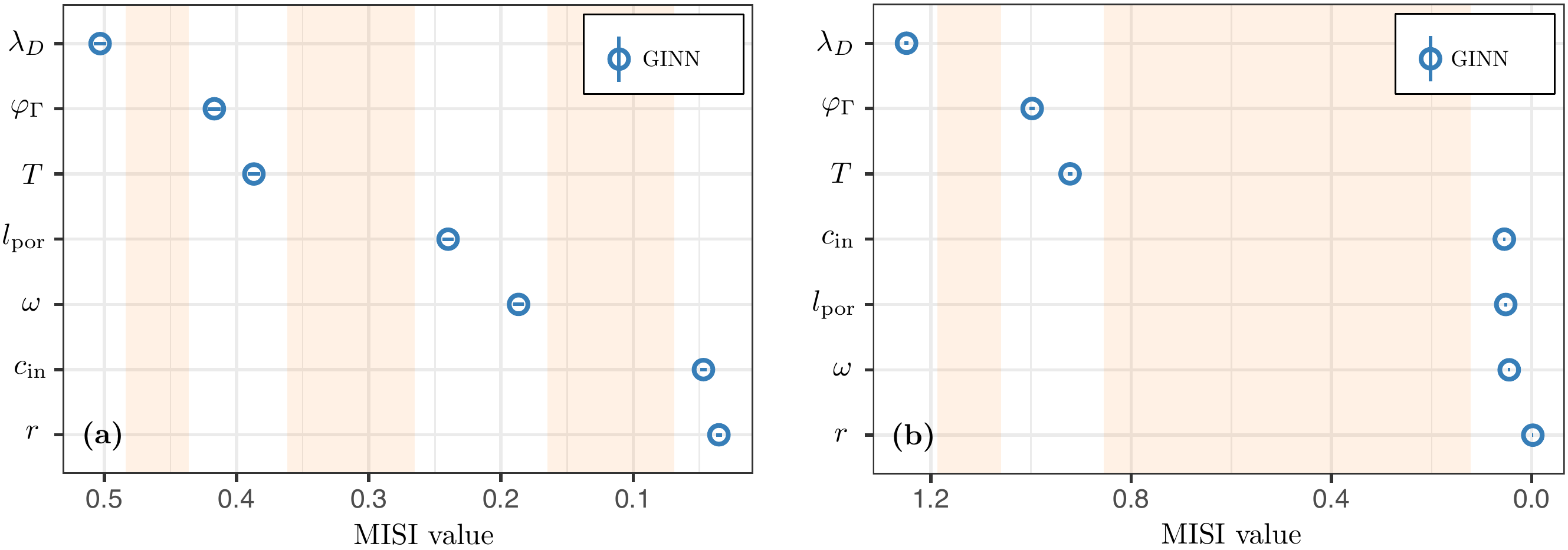}
  \caption{Plug-in Monte Carlo estimators of the first-order MISI
    rankings in the large $\keff$ regime. The latter corresponds to
     the reduced ranges of (a) $\lambda_D$, $\omega$ and (b)
     $\lambda_D$, $\lpor$ indicated by the dashed boxes in \cref{fig:close_design_loop}a and b, respectively. These new rankings are based on $\num{5e4}$ samples generated by the GINN surrogate. They differ from those in
    \cref{fig:ranking-1st} and inform subsequent decision
    tasks in the multiscale design process.}
  \label{fig:close_design_loop_rank}
\end{figure}

\Cref{fig:close_design_loop}a,b shows a clear gradient in the response surfaces, %
with $\keff$ being largest when $\lambda_D$ is small and either $\omega$ or $\lpor$ is large; this observation follows directly from~\cref{eq:lambda_D} and \eqref{eq:kappa-eff}. For the former case, we  zoom in on the region $T\in[208,360]~\si{\kelvin}$ and $\cin\in[0.9,1.08]~\si{\mol\per\liter}$, such that $\lambda_D\in[0.0771,0.1109]~\si{\nano\meter}$ and $\omega\in[0.7,0.8375]$. For the latter case, we consider the region $T\in[208,350]~\si{\kelvin}$ and $\cin\in[0.95,1.08]~\si{\mol\per\liter}$ (such that $\lambda_D\in[0.0771,0.1065]~\si{\nano\meter}$) and $r\in[1.5,1.75]~\si{\nano\meter}$ and $\omega\in[0.79,0.8375]$ (such that $\lpor\in[1.4030,2.0965]~\si{\nano\meter}$). \Cref{fig:close_design_loop_rank} visualizes the new first-order MISI rankings in the reduced parameter space suggested by the most relevant parameter directions in \cref{fig:close_design_loop}a,b that were informed by the first-order (\cref{fig:ranking-1st}) and second-order (\cref{fig:ranking-2nd}) effect rankings. While $\lambda_D$ still has the biggest impact, $\phiG$ and $T$ are now the second- and third-most important CVs in both cases, while $r$ remains the least important CV. Repeating this process informs subsequent decision tasks, yielding a procedure to iteratively refine the materials design. A similar reasoning can be followed based on the response surface of $\keff$ for variations in $\omega$ and $\lpor$ in Fig.~\ref{fig:close_design_loop}c. The narrower shape of this surface reflects the correlation between these CVs, in accordance with~\cref{eq:lpor}.

Comparison of the first-order rankings in \cref{fig:ranking-1st} with the second-order rankings in \cref{fig:ranking-2nd} reveals that even though the individual contributions of $\phiG$ and $T$ to changes in $\keff$ are smaller than those of $\lambda_D$, $\lpor$ and $\omega$, the effect of their interactions dominates that of the pairwise interactions between the latter. This can be explained by the different degrees of dependence among these CVs: $\lambda_D$ is not directly related to $\lpor$ or $\omega$, while the pairs $(T,\phiG)$ and $(\omega,\lpor)$ are combinations of CVs that depend on each other through physical relations (see \cref{eq:phiG} and \cref{eq:lpor}, respectively). This is reflected by the various shapes of the response surfaces in \cref{fig:close_design_loop}. This result enables one to reduce parameter ranges for $\phiG$ and $T$ to close design loops and demonstrates the importance of the higher-order effects (i.e., due to interactions between CVs) introduced in \cref{sec:high-order-effects} for design loop closure in complex systems.

Since the new parameter ranges considered for predicting the effect rankings in \cref{fig:close_design_loop_rank} lie inside the original parameter space, we used the existing trained GINN to predict the new response samples. Alternatively, the MISI rankings might suggest the exploration of new parameter ranges outside the original space, thereby guiding the generation of (a limited amount of) new physics-based data corresponding to those ranges and retraining the GINN with this new training dataset. For deep neural networks with many hidden layers, transfer learning \cite{Pratt:1993} can be employed to reuse most of the existing network and thereby reduce the cost of training the new surrogate.

\section{Conclusions and Outlook}
\label{sec:conclusions}

We developed a moment-independent global sensitivity analysis (GSA) based on differential mutual information (MI). Mutual Information Sensitivity Indices (MISIs) provide a model-agnostic mechanism for ranking the impact of correlated tunable control variables (CVs) on quantities of interest (QoIs). The high computational cost of querying physics-based models, typically a barrier to the use of such data-driven methods, is ameliorated by leveraging deep learning-based surrogate models that enable fast  generation of response data.
Although these black boxes do not generate engineering insights, our MI-based GSA  allows one to \ADD{interrogate} surrogates and \ADD{explain surrogate predictions in the context of the physics-based model to close engineering design loops}. %

We utilized our MI-based GSA in conjunction with a recently developed Graph-Informed Neural Network (GINN) \cite{HallTaverniersEtAl:2020aa}, capable of handling correlated CVs, and tested our approach on \ADD{two models of interest in energy storage including} a multiscale model of an electrical double-layer capacitor. 
We presented two algorithms for estimating and ranking MISIs and, for the applications of interest, calculated first order MISIs to capture individual CV effects and second-order MISIs to capture the effects of pairwise interactions between CVs.
We validated the first-order MISI rankings against both physics- and GINN-based QoI response surfaces. Finally, we closed the engineering design loop by considering the most sensitive input directions and investigating effect rankings in a reduced parameter space corresponding to large effective conductivity values.

Our analysis leads to the following major conclusions.
\begin{enumerate}
\item \ADD{Viewed as an uncertainty quantification for surrogate models,} our MI-based GSA works seamlessly with the GINN---a promising result that encourages its application to interrogate other black-box surrogate models.
\item The GINN enables generation of a predicted dataset whose size is one-to-two orders of magnitude larger than that of the training set provided by the physics-based model. That translates into well-resolved first-order MISI rankings facilitated by either comparison-adjusted (Algorithm 1) or percentile (Algorithm 2) confidence intervals. At a comparable computational cost, the original physics-based model cannot distinguish these rankings with confidence.
\item The resolved rankings produced with the GINN are consistent between Algorithms 1 and 2.
\item The most/least sensitive CVs identified through the first-order MISI effect rankings show a nonlinear/nearly flat response curve for the QoIs, supporting the validity of the MI-based approach. The relative magnitudes of the largest first-order MISIs for both QoIs produced by Algorithm 1 are also in line with the differences between the normalized response curves; this holds true for the smallest first-order MISIs as well.
\item The impact of mutual interactions between correlated CVs on the QoIs needs to be taken into account via higher-order MISIs. Pairwise interactions between CVs with small individual contributions to the QoIs can dominate those between CVs with larger additive effects.
\item Within a reduced parameter space leading to optimal QoI values, the relative importance of the various CVs is different from that in the original parameter space (that could be nonlinear, see \cref{fig:close_design_loop}c,d). This new ranking informs subsequent design cycles, spawning an iterative procedure that enables rapid prototyping and reduces time to market.
\end{enumerate}
Motivated by its successful application to GINNs, we aim to pair our MI-based GSA with other deep neural network surrogates including physics-informed and physics-constrained neural networks. In particular, our MISI rankings could help simplify the custom loss functions of those deep neural networks by \ADD{filtering} out less important parameters.

\section{Acknowledgments}
 
The research of S.~T. and D.~T. was partially supported by the Air Force Office of Scientific Research (AFOSR) under grant FA9550-18-1-0474 and by a gift from Total, both awarded to D.~T. A portion of this research was undertaken when E.~H. was a postdoctoral research scientist in the Chair of Mathematics for Uncertainty Quantification at RWTH Aachen University, Germany and was partially supported by the Alexander von Humboldt Foundation. The research of M.~K. was partially supported by  the Air Force Office of Scientific Research (AFOSR) under grant FA-9550-18-1-0214 and by the National Science Foundation (NSF) under grants  DMS-2008970  and  CISE-1934846.

\bibliographystyle{elsarticle-num} \bibliography{arxivsubmit}%

\end{document}